# Ranking-Based Black-Box Complexity[*]

Benjamin Doerr    Carola Winzen[†]

Max Planck Institute for Informatics, 66123 Saarbrücken, Germany
{doerr|winzen}@mpi-inf.mpg.de

**Abstract**

Randomized search heuristics such as evolutionary algorithms, simulated annealing, and ant colony optimization are a broadly used class of general-purpose algorithms. Analyzing them via classical methods of theoretical computer science is a growing field. While several strong runtime analysis results have appeared in the last 20 years, a powerful complexity theory for such algorithms is yet to be developed. We enrich the existing notions of black-box complexity by the additional restriction that not the actual objective values, but only the relative quality of the previously evaluated solutions may be taken into account by the black-box algorithm. Many randomized search heuristics belong to this class of algorithms.

We show that the new ranking-based model can give more realistic complexity estimates. The class of all binary-value functions has a black-box complexity of $O(\log n)$ in the previous black-box models, but has a ranking-based complexity of $\Theta(n)$.

On the other hand, for the class of all OneMax functions, we present a ranking-based black-box algorithm that has a runtime of $\Theta(n/\log n)$, which shows that the OneMax problem does not become harder with the additional ranking-basedness restriction.

**Keywords:** Query complexity; theory of randomized search heuristics; Mastermind; black-box complexity.

## 1 Introduction

Randomized search heuristics are general purpose algorithms for optimization problems. They include bio-inspired approaches such as evolutionary algorithms and ant colony optimization, but also classical approaches like random search or randomized hill-climbers.

In practice, randomized search heuristics often are highly successful and thus extensively used [19]. They have the additional advantage that not too much understanding of the optimization problem at hand is needed, and that once implemented, they can easily be re-used for similar problems.

One of the difficulties in using such heuristics is that it is very hard to predict which problems are easy for a suitable heuristic and which are generally intractable for randomized search heuristics. In contrast to a large body of empirical work on this problem, there has been much less theoretical work. This work mostly lead to results for particular problems and particular heuristics. Droste, Jansen, and Wegener [9] determined the runtime of the $(1+1)$ evolutionary algorithm (EA) for several important test function classes. Another example is the work by

---

[*]A preliminary version of the results appeared in [7].

[†]Corresponding author. Full contact details: Max Planck Institute for Informatics, D1: Algorithms and Complexity, Campus E 1 4, 66123 Saarbrücken, Germany. Mail: winzen@mpi-inf.mpg.de. Phone: +49 681 9325 1013. Fax: +49 681 9325 1099.

Neumann and Wegener [21], which shows that the $(1 + 1)$ EA finds a minimum spanning tree using $O(m^2 \log m)$ function evaluations in connected graphs having $m$ edges and polynomially bounded edge weights.

Still, for a broader understanding of what are easy and difficult problems, a complexity theory similar to what exists in classical algorithmics would be highly desirable also for randomized search heuristics. The seminal paper by Droste, Jansen, and Wegener [10], introducing the so-called *unrestricted black-box model*, appears to be the first attempt to start such a complexity theory in the randomized search heuristics community.

The paradigm that randomized search heuristics should ideally be problem-independent implies that the only way a randomized search heuristic can obtain problem-specific information is by evaluating solution candidates. This evaluation is done by an oracle that returns objective values, but reveals no further information about the objective function. An algorithm that has no access to the objective function (and thus has no access to the optimization problem to be solved) other than by querying objective values from such an oracle, is called a *black-box algorithm*.

Given a class of functions $\mathcal{F}$, Droste et al. define the unrestricted black-box complexity of $\mathcal{F}$ to be the minimum (taken over all black-box algorithms) expected number of function evaluations needed to optimize any function $f \in \mathcal{F}$. This number, naturally, is a lower bound on the runtime of any randomized search heuristic for the class $\mathcal{F}$.

In classical theoretical computer science, unrestricted black-box complexity is also studied under the notion of *randomized query complexity*. Many results for a variety of problems exist. Out of the many examples let us mention the problem of finding a local minimum of an unknown pseudo-Boolean function $f : \{0, 1\}^n \to \mathbb{R}$. This problem has been studied intensively in the computer science literature, for deterministic algorithms (deterministic black-box complexity; cf., e.g., the work by Llewellyn, Tovey, and Trick [18]) and for randomized algorithms, for example by Aldous [2], Aaronson [1], and by Zhang [25]. Zhang gives a tight $\Theta(2^{n/2}\sqrt{n})$ bound for the randomized query complexity of finding a local minimum.

Originally motivated by the coin-weighing problem, a much earlier work studying the unrestricted black-box complexity of the generalized ONEMAX function class (definition follows) is the one by Erdős and Rényi [12], cf. Section 4. This ONEMAX function class is also strongly related to the well-known Mastermind game, a game that has gained much attention from the computer science community. For example, Chvátal [4] studies a general version of this game with $k$ colors and $n$ positions. That is, the secret code is a length-$n$ string $z \in \{0, 1, \ldots, k-1\}^n$. Chvátal shows that for any constant number $k$ of colors the codebreaker has a strategy revealing the secret code using only $\Theta(n/\log n)$ guesses. In our notation this result is equivalent to saying that the unrestricted black-box complexity of the generalized ONEMAX function class is $\Theta(n/\log n)$. The connection between unrestricted black-box complexity and randomized query complexity was seemingly overlooked so far in the randomized search heuristics community. Similarly, it seems that the community was not aware of the existing results for the Mastermind game.

Unfortunately, it turned out that regarding all black-box algorithms leads to sometimes unexpectedly small complexities (obtained by not very sensible algorithms). As a trivial example, note that the unrestricted black-box complexity of any class of functions $\mathcal{F} = \{f\}$ consisting of a single objective function is one. This is certified by the black-box algorithm that simply queries the optimum of $f$.

This and further examples suggest that a restriction of the class of algorithms regarded might lead to more meaningful results. A major step in this direction is the work by Lehre and Witt [17]. They introduce a so-called *unbiased black-box model*, which, among other restrictions to the class of algorithms regarded, requires that all search points queried by the algorithm must be obtained from previous or random search points by so-called *unbiased variation operators* (see Section 2 for the full details). When only unary operators are allowed, this leads to a lower bound of $\Omega(n \log n)$ for the complexity of any single-element class $\mathcal{F} = \{f\}$ with $f$ having a unique global optimum. This is, indeed, the typical runtime of simple search heuristics like randomized hill-climbers on simple function classes like strictly monotone functions; i.e., functions $f : \{0,1\}^n \to \mathbb{R}$ with $f(x) < f(y)$ for all $x, y \in \{0,1\}^n$ for which (a) $x_i = 0$ implies $y_i = 0$, $1 \le i \le n$ and (b) there is at least one index $i$ such that $x_i = 1 = 1 - y_i$.

In this work, we shall argue that the unbiased model of Lehre and Witt is still not restrictive enough, and we will propose an alternative model. Let $f : \{0,1\}^n \to \mathbb{R}, x \mapsto \sum_{i=1}^{n} 2^{i-1} x_i$ be the binary-value function of the bit string $x$. Let $\text{BinaryValue}_n^*$ be the the class of functions consisting of $f$ and all functions obtained from $f$ by permuting the order of the bit positions and by flipping the meaning of the values of some bit-positions. In other words, $\text{BinaryValue}_n^*$ is the smallest class of functions containing $f$ that is invariant under first applying an automorphism of the discrete $n$-dimensional hypercube $\{0,1\}^n$. Then, as we shall show in this paper, the unbiased black-box complexity of $\text{BinaryValue}_n^*$ is at most $\lceil \log_2 n \rceil + 2$, if we allow the variation operators to be of arbitrary arity. The corresponding black-box algorithm (see Section 5) heavily exploits knowing the precise objective values of queried search points.

This is what most randomized search heuristics do not do. They typically only use the objective values to *compare* search points. We define a black-box complexity notion referring to this paradigm by allowing the algorithms to only exploit the relative order of the search points queried so far. In other words, throughout the optimization process the black-box algorithm knows for any two search points $x$ and $y$ queried so far only whether $f(x) < f(y)$, $f(x) = f(y)$, or $f(x) > f(y)$. In particular, it does not know the true values of $f(x)$ and $f(y)$. This model captures many commonly used randomized search heuristics, e.g., many evolutionary algorithms, hill climbers like Random Local Search, and ant colony optimization.

We show that our *ranking-based black-box model* overcomes some drawbacks of the previous models. For example, for the binary-value function class $\text{BinaryValue}_n^*$ introduced above, both the otherwise unrestricted ranking-based black-box complexity and the unbiased ranking-based black-box complexity are of order $\Theta(n)$ instead of $O(\log n)$ without the ranking restriction. In the $\Theta(n)$ statement, the lower bound is clearly the more interesting one. This bound holds already for the subclass $\text{BinaryValue}_n$ of $\text{BinaryValue}_n^*$ consisting of all functions

$$f_z : \{0,1\}^n \to \mathbb{R}, x \mapsto \sum_{i=1}^{n} 2^{i-1} (x_i \oplus z_i),$$

$z \in \{0,1\}^n$.

The upper bound is easily verified by a simple hill-climber that, in arbitrary order, changes a single bit-value of the current solution and accepts the new solution if it is better than the previous one. In summary, we see that for this function class, the ranking-based black-box complexity seems to give us a more natural complexity measure than the previous approaches.

We also analyze the ranking-based black-box complexity of a second function class that is often regarded in theoretical analyses of randomized search heuristics, namely the $\text{OneMax}$ function class. Let $\text{OneMax}_n$ be the class of all so-called $\text{OneMax}$ functions

$$f_z : \{0,1\}^n \to \mathbb{R}, x \mapsto |\{i \in [0,n] \cap \mathbb{Z} \mid x_i = z_i\}|,$$

$z \in \{0,1\}^n$. Hence, $f_z(x)$ is the number of bit positions in which $x$ and $z$ coincide. Here, both the unrestricted and the unbiased black-box complexity are $\Theta(n/\log n)$, which is slightly smaller than the $\Theta(n \log n)$ needed by most randomized search heuristics. The proofs of the $O(n/\log n)$ black-box complexity results again heavily exploit that the oracle returns the precise objective values ("fitness values"). They all build on the beautiful idea that $\Theta(n/\log n)$ random search points together with their (precise) fitness determine the hidden objective function, cf. [3, 5, 12]. In spite of this, we present a ranking-based black-box algorithm that still solves the problem with $\Theta(n/\log n)$ queries.

Our result on the ONEMAX class can also be interpreted in the context of the Mastermind game. As many authors (cf. [4, 15]) we regard the black-peg version of the game, where—instead of answering with black and white answer-pegs indicating in how many positions the secret code of the codemaker and the guess of the codebreaker coincide (black answer-pegs) and how many additional colors are correct but in the wrong position (white answer-pegs)—the codemaker does only reply with black answer-pegs. The ranking-based black-box complexity of ONEMAX$_n$ corresponds to the black-peg version of the Mastermind game in which the codemaker does respond to the codebreaker's guesses by providing a ranking of the guesses queried so far. This ranking is based on the number of black answer-pegs only. Then, as in the original generalized Mastermind game with black and white answer-pegs, the codebreaker still has an optimal winning strategy using only $\Theta(n/\log n)$ guesses.

These two results show that in some cases, the additional restriction that only relative qualities of solutions may be taken into account does give more insightful problem difficulty estimates, whereas in other cases the ranking restriction does not change the unexpectedly low difficulty estimates given by the previous black-box models.

We should note that there are two related research works in the literature. In [8, 10], for certain classes $\mathcal{F}$ of functions the unrestricted black-box complexity of $\{h \circ f \mid f \in \mathcal{F}, h : \mathbb{R} \to \mathbb{R}$ strictly monotonically increasing$\}$ is regarded. This is closely related (see Section 3) to the ranking-based black-box complexity, as introduced here, of the class $\mathcal{F}$. The connection to black-box algorithms not exploiting the absolute function values, however, is not made there.

Teytaud and co-authors [13, 23] give general lower bounds for the convergence rate of comparison-based and ranking-based evolutionary strategies in continuous domains. From these works results for discrete domains can be obtained. We do not see, however, that for such domains their lower bounds are stronger than the natural information-theoretic ones which were already observed in [8, 10].

## 2 Preliminaries and Previous Black-Box Models

In this section, we give a brief overview of two previous black-box models, the *unrestricted black-box model* by Droste, Jansen, and Wegener [10] and the more recent *unbiased black-box model* by Lehre and Witt [17].

Let us first fix the notations used frequently throughout the paper.

### 2.1 Notation

The positive integers are denoted by $\mathbb{N}$. For $k \in \mathbb{N}$, we abbreviate $[k] := \{1, \ldots, k\}$. Similarly, we define $[0..k] := [k] \cup \{0\}$. For $k, \ell \in \mathbb{N}$ we write $[k \pm \ell] := [k - \ell, k + \ell] \cap \mathbb{Z}$.

Let $n \in \mathbb{N}$. For a bit string $x = x_1 \ldots x_n \in \{0, 1\}^n$, we denote by $\bar{x}$ the bitwise complement of $x$, i.e., for all $j \in [n]$ we have $\bar{x}_j = 1 - x_j$.

For $n \in \mathbb{N}$ and $j \in [n]$ by $e_j^n$ we denote the $j$-th unit vector of length $n$.

If $x, y \in \{0, 1\}^n$, we obtain the bit string $x \oplus y$ by setting, for each $j \in [n]$, $(x \oplus y)_i := 1$ if $x_i \neq y_i$, and $(x \oplus y)_i := 0$ if $x_i = y_i$. That is, $\oplus$ denotes the bitwise exclusive-or. We use the shorthand $|x|_1$ for the number of ones in the bit string $x$, i.e., $|x|_1 = \sum_{i=1}^n x_i$.

If $f$ is a function and $S$ a set, we write $f(S) := \{f(s) \mid s \in S\}$. We write $\text{id}_S$ for the identity function of $S$, i.e., $\text{id}_S(s) = s$ for all $s \in S$. For $n \in \mathbb{N}$, the set $S_n$ contains all permutations of $[n]$. For $\sigma \in S_n$ and $x \in \{0,1\}^n$ we abbreviate $\sigma(x) := x_{\sigma(1)} \ldots x_{\sigma(n)}$.

For two real values $a, b \in \mathbb{R}$ with $a > b$ the interval $[a, b]$ is defined to be the empty set.

Lastly, we denote by ln the natural logarithm to base $e := \exp(1)$. If we refer to a logarithm of a different base, we indicate this in the subscript; e.g., we write $\log_2$ for the binary logarithm.

All asymptotic notation (Landau symbols, big-Oh notation) will be with respect to $n$, which typically denotes the dimension of the search space $\{0,1\}^n$.

## 2.2 Useful Tools

Throughout the paper we shall apply several versions of Chernoff's bound. The following can be found, e.g., in [11].

**Lemma 1** (Chernoff bounds). *Let $X = \sum_{i=1}^n X_i$ be the sum of $n$ independently distributed random variables $X_i$, where each variable $X_i$ takes values in $[0,1]$. Then the following statements hold.*

$$\forall t > 0 : \Pr[X > \mathrm{E}[X] + t] \leq \exp(-2t^2/n) \text{ and } \Pr[X < \mathrm{E}[X] - t] \leq \exp(-2t^2/n), \quad (1)$$
$$\forall \varepsilon > 0 : \Pr\left[X < (1-\varepsilon)\mathrm{E}[X]\right] \leq \exp(-\varepsilon^2 \mathrm{E}[X]/2). \quad (2)$$

We shall also use the following estimate on factorials. It is a direct consequence of Stirling's formula. The version presented below is due to Robbins [22].

**Lemma 2** (factorials). *For all $n \in \mathbb{N}$,*

$$\sqrt{2\pi} n^{n+1/2} e^{-n} e^{1/(12n+1)} \leq n! \leq \sqrt{2\pi} n^{n+1/2} e^{-n} e^{1/(12n)}.$$

## 2.3 Unrestricted and Unbiased Black-Box Complexity

Usually, the complexity of a problem is measured by the performance of the best algorithm out of some class of algorithms, e.g., all algorithms which can be implemented on a Turing machine [14, 16].

What distinguishes randomized search heuristics from classical algorithms is that they are problem-independent. As such, the only way they obtain information about the problem to be solved is by learning the objective value of possible solutions ("search points"). To ensure problem-independence, one usually assumes that the objective function is given by an oracle or as a *black-box*. Using this oracle, the algorithm may query the objective value of any search point. Such a query returns the objective value of the search point, but no other information about the objective function.

For simplicity, we shall restrict ourselves to real-valued objective functions defined on the set $\{0,1\}^n$ of bit strings of length $n$. This is motivated by the fact that many evolutionary algorithms use such a representation.

Naturally, we do allow that the algorithms use random decisions. From the black-box concept, it follows that the only type of action the algorithm may perform is, based on the objective values learned so far, to decide on a probability distribution over $\{0,1\}^n$, to sample a search point $x \in \{0,1\}^n$ according to this distribution, and to query its objective value from the oracle. This leads to the scheme of Algorithm 1, which we call an *unrestricted black-box algorithm*.

In typical applications of randomized search heuristics, evaluating the fitness of a search point is more costly than the generation of a new search point. For this reason we take as performance measure of a black-box algorithm the number of queries to the oracle until an optimal

**Algorithm 1:** Scheme of an unrestricted black-box algorithm

1 **Initialization:** Sample $x^{(0)}$ according to some probability distribution $p^{(0)}$ over $\{0,1\}^n$ and query $f(x^{(0)})$;
2 **Optimization: for** $t = 1, 2, 3, \ldots$ **do**
3   Depending on $\big((x^{(0)}, f(x^{(0)})), \ldots, (x^{(t-1)}, f(x^{(t-1)}))\big)$ choose a probability distribution $p^{(t)}$ over $\{0,1\}^n$ and sample $x^{(t)}$ according to $p^{(t)}$;
4   Query $f(x^{(t)})$;

search point is queried for the first time. Since we are interested in randomized algorithms, we regard the expected number of queries.

Formally, for an unrestricted algorithm $A$ and a function $f : \{0,1\}^n \to \mathbb{R}$, let $T(A, f) \in \mathbb{R} \cup \{\infty\}$ be the expected number of fitness evaluations until $A$ queries for the first time some $x \in \arg\max f$. We call $T(A, f)$ the *runtime of $A$ for $f$* or, likewise, the *optimization time of $A$ for $f$*. We can now follow the usual approach in complexity theory. For a class $\mathcal{F}$ of functions $\{0,1\}^n \to \mathbb{R}$, the *$A$-black-box complexity of $\mathcal{F}$* is $T(A, \mathcal{F}) := \sup_{f \in \mathcal{F}} T(A, f)$, the worst-case runtime of $A$ on $\mathcal{F}$. Let $\mathcal{A}$ be a class of black-box algorithms for functions $\mathcal{F}$. Then the *$\mathcal{A}$-black-box complexity of $\mathcal{F}$* is $T(\mathcal{A}, \mathcal{F}) := \inf_{A \in \mathcal{A}} T(A, \mathcal{F})$, the minimum ("best") complexity among all $A \in \mathcal{A}$ for $\mathcal{F}$. If $\mathcal{A}$ is the class of all unrestricted black-box algorithms, we call $T(\mathcal{A}, \mathcal{F})$ the *unrestricted black-box complexity* of $\mathcal{F}$. This is the black-box complexity as introduced by Droste, Jansen, and Wegener [10].

As mentioned in the introduction, it is easily seen that the class of all unrestricted black-box algorithms is very powerful. For example, for any function class $\mathcal{F} = \{f\}$ consisting of one single function, the unrestricted black-box complexity of $\mathcal{F}$ is 1. The algorithm that simply queries an optimal solution of $f$ as first action shows this bound.

This and related drawbacks of the unrestricted black-box model inspired Lehre and Witt [17] to introduce a more restrictive black-box model, where algorithms may generate new solution candidates only from random or previously generated search points and only by using unbiased operators. Still this model contains most of the commonly studied search heuristics, such as many $(\mu+\lambda)$ and $(\mu,\lambda)$ evolutionary algorithms, simulated annealing, the Metropolis algorithm, and Random Local Search.

**Definition 3** (*k*-ary unbiased variation operator). *Let $k \in \mathbb{N}$. A $k$-ary unbiased distribution $(D(. \mid y^{(1)}, \ldots, y^{(k)}))_{y^{(1)}, \ldots, y^{(k)} \in \{0,1\}^n}$ is a family of probability distributions over $\{0,1\}^n$ such that for all inputs $y^{(1)}, \ldots, y^{(k)} \in \{0,1\}^n$ the following two conditions hold.*

$(i) \; \forall x, z \in \{0,1\}^n : D(x \mid y^{(1)}, \ldots, y^{(k)}) = D(x \oplus z \mid y^{(1)} \oplus z, \ldots, y^{(k)} \oplus z),$

$(ii) \; \forall x \in \{0,1\}^n \; \forall \sigma \in S_n : D(x \mid y^{(1)}, \ldots, y^{(k)}) = D(\sigma(x) \mid \sigma(y^{(1)}), \ldots, \sigma(y^{(k)})) \,.$

*We refer to the first condition as $\oplus$-invariance and to the second as permutation invariance. A variation operator creating an offspring by sampling from a $k$-ary unbiased distribution is called a $k$-ary unbiased variation operator.*

Note that the only 0-ary unbiased distribution over $\{0,1\}^n$ is the uniform one. 1-ary operators, also called *unary* operators, are sometimes referred to as mutation operators, in particular in the field of evolutionary computation. 2-ary operators, also called *binary* operators, are often referred to as crossover operators. If we allow arbitrary arities, we call the corresponding black-box model the $*$-ary unbiased black-box model.

$k$-ary unbiased black-box algorithms can now be described via the scheme of Algorithm 2. The *$k$-ary unbiased black-box complexity* of some class of functions $\mathcal{F}$ is the complexity of $\mathcal{F}$ with respect to all $k$-ary unbiased black-box algorithms.

---

**Algorithm 2:** Scheme of a $k$-ary unbiased black-box algorithm

1 **Initialization:** Sample $x^{(0)} \in \{0,1\}^n$ uniformly at random and query $f(x^{(0)})$;
2 **Optimization:** for $t = 1, 2, 3, \ldots$ do
3     Depending on $\left(f(x^{(0)}), \ldots, f(x^{(t-1)})\right)$ choose $k$ indices $i_1, \ldots, i_k \in [0..t-1]$ and a $k$-ary unbiased distribution $(D(. \mid y^{(1)}, \ldots, y^{(k)}))_{y^{(1)}, \ldots, y^{(k)} \in \{0,1\}^n}$;
4     Sample $x^{(t)}$ according to $D(. \mid x^{(i_1)}, \ldots, x^{(i_k)})$ and query $f(x^{(t)})$;

---

As we mentioned in the introduction, Lehre and Witt [17] proved, among other results, that all functions with a single global optimum have a unary unbiased black-box complexity of $\Omega(n \log n)$. For several standard test problems this bound is met by classical unary randomized search heuristics such as the $(1+1)$ evolutionary algorithm or Random Local Search. Recall that, as pointed out above, the unrestricted black-box complexity of any such function is 1.

## 3 The Ranking-Based Black-Box Model

Since many standard randomized search heuristics do not take advantage of knowing the *exact* objective values but rather take into account only the relative quality of search points, Nikolaus Hansen (INRIA Saclay, France; personal communication) suggested to develop a corresponding black-box model. In fact, many heuristics create new search points based only on how the objective values of the previously queried search points compare. That is, after having queried $t$ fitness values $f(x^{(0)}), \ldots, f(x^{(t-1)})$, they rank the corresponding search points $x^{(0)}, \ldots, x^{(t-1)}$ according to their relative fitness. The selection of input individuals $x^{(i_1)}, \ldots, x^{(i_k)}$ for the next variation operator is based solely on this ranking.

We define the ranking induced by $f$ as follows.

**Definition 4** (ranking induced by $f$)**.** *Let $S$ be a set, let $f : S \to \mathbb{R}$ be a function, and let $C$ be a finite subset of $S$. The* ranking $\rho_C$ *of $C$ induced by $f$ assigns to each element $c \in C$ the number of elements in $C$ with a smaller $f$-value plus 1, formally, $\rho_C(c) := 1 + |\{c' \in C \mid f(c') < f(c)\}|$.*

Note that two elements with the same $f$-value are assigned the same rank.

As discussed above, when selecting a parent population for generating new search points, many randomized search heuristics do only use the ranking of the search points seen so far. In this work we are interested in how this fact influences the complexity of standard test function classes. Therefore, we regard here the restricted class of black-box algorithms that use no other information than this ranking. This yields the following black-box models.

The *unrestricted ranking-based black-box complexity* of some class of functions is the complexity with respect to all algorithms following the scheme of Algorithm 3.

Similarly, the *$k$-ary unbiased ranking-based black-box complexity* of some class of functions is the complexity with respect to all algorithms following the scheme of Algorithm 4.

Both ranking-based black-box models capture many common search heuristics such as evolutionary algorithms using elitist selection, ant colony optimization, and Random Local Search. They do not include algorithms like simulated annealing, threshold accepting, evolutionary algorithms using fitness-proportional selection, or the Metropolis algorithm.

**Algorithm 3:** Scheme of an unrestricted ranking-based black-box algorithm

1 **Initialization:** Sample $x^{(0)}$ according to some probability distribution $p^{(0)}$ over $\{0,1\}^n$ and query $f(x^{(0)})$;
2 **Optimization: for** $t = 1, 2, 3, \ldots$ **do**
3      Depending on the ranking of $\{x^{(0)}, \ldots, x^{(t-1)}\}$ induced by $f$, choose a probability distribution $p^{(t)}$ over $\{0,1\}^n$ and sample $x^{(t)}$ according to $p^{(t)}$;
4      Query the ranking of $\{x^{(0)}, \ldots, x^{(t)}\}$ induced by $f$;

**Algorithm 4:** Scheme of a $k$-ary unbiased ranking-based black-box algorithm

1 **Initialization:** Sample $x^{(0)} \in \{0,1\}^n$ uniformly at random and query $f(x^{(0)})$;
2 **Optimization: for** $t = 1, 2, 3, \ldots$ **do**
3      Depending on the ranking of $\{x^{(0)} \ldots, x^{(t-1)}\}$ induced by $f$, choose $k$ indices $i_1, \ldots, i_k \in [0..t-1]$ and a $k$-ary unbiased distribution $(D(. \mid y^{(1)}, \ldots, y^{(k)}))_{y^{(1)}, \ldots, y^{(k)} \in \{0,1\}^n}$;
4      Sample $x^{(t)}$ according to $D(. \mid x^{(i_1)}, \ldots, x^{(i_k)})$ and query the ranking of $\{x^{(0)}, \ldots, x^{(t)}\}$ induced by $f$;

To distinguish the unrestricted and the unbiased black-box model from their ranking-based counterparts, we shall sometimes refer to them as the *basic* unrestricted black-box model and the *basic* unbiased black-box model, respectively.

When working with the ranking-based models, the fact that the rank of a search point $x$ varies with the number of already queried search points may be distracting. However, the ranking-based models can be equivalently modeled via an unknown adaptive *monotone perturbation* of the fitness values. By this we mean that instead of ranking all previously queried search points, the oracle may as well reply to any query $x^{(t)}$ with some value $g(f(x^{(t)}))$, where $f$ is the secret function to be optimized and $g : \mathbb{R} \to \mathbb{R}$ is a strictly monotone function that depends on all search points $x^{(1)}, \ldots, x^{(t)}$ queried so far.

To make this model more precise, let us first recall that a function is said to be *strictly monotone* if for all $\alpha < \beta$ we have $g(\alpha) < g(\beta)$. We show how the oracle can construct such a strictly monotone function "on the fly", preserving the ranking of the search points. Let $A$ be a black-box algorithm. When algorithm $A$ queries a search point $x^{(0)}$ for initialization, the oracle responds to $A$ with "$(g \circ f)(x^{(0)}) = 0$". That is, it sets $g(f(x^{(0)})) := 0$. For any iteration $t$, if algorithm $A$ queries $x^{(t)}$, the oracle returns to $A$ the value

$g(f(x^{(t)})) =$

$$\begin{cases} g(f(x^{(i)})), & \text{if } \rho(x^{(t)}) = \rho(x^{(i)}) \text{ for some } i \in [0..t-1], \\ \max\{g(f(x^{(i)})) \mid i \in [0..t-1]\} + 2^n, & \text{if } \rho(x^{(t)}) = 1, \\ \min\{g(f(x^{(i)})) \mid i \in [0..t-1]\} - 2^n, & \text{if } \rho(x^{(t)}) = t+1, \\ \left(g(f(x^{(h)})) - |g(f(x^{(i)}))|\right)/2, & \text{if } \rho(x^{(h)}) = \max\{\rho(x^{(\ell)}) \mid \rho(x^{(\ell)}) < \rho(x^{(t)})\} \\ & \text{and } \rho(x^{(i)}) = \min\{\rho(x^{(\ell)}) \mid \rho(x^{(\ell)}) > \rho(x^{(t)})\}, \end{cases}$$

where we abbreviate $\rho := \rho_{\{x^{(i)} \mid i \in [0..t]\}}$, the ranking of $\{x^{(i)} \mid i \in [0..t]\}$ induced by $f$. It is easily verified that indeed we have $g(f(x^{(i)})) > g(f(x^{(j)}))$ if and only if $f(x^{(i)}) > f(x^{(j)})$, i.e., $g$ can be extended to a strictly monotone function $\mathbb{R} \to \mathbb{R}$.

As any stage of the run of the black-box algorithm, the information revealed by the $(g \circ f)$-values and the information revealed by the ranking of the search points is the same. Therefore, the two models are equivalent.

We sometimes refer to the model with a $(g \circ f)$-oracle as the (unrestricted or unbiased, respectively) *monotone black-box model*. In particular for proving upper bounds this model is more convenient to work with.

**Convention.** In what follows, we shall always denote by $g$ the monotone perturbation. That is, $g$ is the function, which is used for representing the ranking of the already queried search points.

In [8, 10], Droste, Jansen, Tinnefeld (only [8]) and Wegener implicitly regard a different notion of black-box complexity without exploiting absolute fitness values. For a given class $\mathcal{F}$ of functions $f : \{0,1\}^n \to \mathbb{R}$, they regard the unrestricted black-box complexity of the monotone closure $\overline{\mathcal{F}} := \{h \circ f \mid f \in \mathcal{F}, h : \mathbb{R} \to \mathbb{R} \text{ strictly monotonically increasing}\}$ of $\mathcal{F}$. Clearly, the optimal search points of $f$ and $h \circ f$ are the same. Moreover, the relative quality of two search points is the same under the fitness function $f$ and $h \circ f$. Hence a black-box algorithm optimizing $h \circ f$, since $h$ is unknown, is in a similar position as a ranking-based black-box algorithm optimizing $f$.

Unfortunately, contrary to the first believe, it is not so obvious whether the unrestricted black-box complexity of $\overline{\mathcal{F}}$ and the ranking-based black-box complexity of $\mathcal{F}$ are the same. This has been informally argued in [8, 10], but we currently do not see a rigorous proof for such a statement. The difficulty is that, theoretically, a black-box algorithm optimizing an unknown, but from that point on fixed, $h \circ f$ might acquire a probabilistic knowledge on $h$ and exploit this in future queries. This might put the algorithm in a better position as in the adaptively monotonically perturbed model described above.

This problem does not exist if we only regard deterministic black-box algorithms. Here we may adaptively change the function $h$ during the optimization progress and argue that in fact we could have started with this $h$. This argument is not possible for randomized algorithms where the distribution of the answers obtained so far has to be taken into account. Likewise, we cannot invoke Yao's minimax principle since $\overline{\mathcal{F}}$ is not finite.

For these reasons, we currently do not know whether the black-box complexity of $\overline{\mathcal{F}}$ and our ranking-based black-box complexity of $\mathcal{F}$ are the same or not. Clearly, the ranking-based black-box complexity is always not less than the black-box complexity of $\overline{\mathcal{F}}$.

Given that we are not sure that both models agree, [8, 10] show that the unrestricted black-box complexity of the monotone closure of BINARYVALUE$_n$ is at least of order $n/\log n$ and at most of order $n + 2$.

## 4 The Ranking-Based Black-Box Complexity of OneMax

A classical easy test function in the theory of randomized search heuristics is the function ONEMAX, which simply counts the number of 1-bits, ONEMAX$(x) = \sum_{i=1}^{n} x_i$. The natural generalization of this particular function to a non-trivial class of functions is defined as follows.

**Definition 5** (ONEMAX function class). *For $z \in \{0,1\}^n$ let*

$$\text{OM}_z : \{0,1\}^n \to [0..n], x \mapsto \text{OM}_z(x) = |\{i \in [n] \mid x_i = z_i\}|.$$

*The string $z = \arg\max \text{OM}_z$ is called the* target string *of $\text{OM}_z$. Let $\text{ONEMAX}_n := \{\text{OM}_z \mid z \in \{0,1\}^n\}$ be the set of all generalized* ONEMAX *functions.*

Our main result is the following.

**Theorem 6.** *The unary unbiased ranking-based black-box complexity of* $\text{ONEMAX}_n$ *is* $\Theta(n \log n)$. *For* $2 \le k \le n$, *the k-ary unbiased ranking-based black-box complexity of* $\text{ONEMAX}_n$ *is* $O(n/\log k)$.

For $k = n^{\Omega(1)}$ this statement is asymptotically optimal since already for the unrestricted black-box complexity a lower bound of $\Omega(n/\log n)$ has been shown by Erdős and Rényi [12] and independently by Chvátal [4] and again later also by Droste, Jansen, and Wegener [10]. Also independently of each other, Erdős and Rényi [12], Chvátal [4], and Anil and Wiegand [3] proved a matching upper bound. This shows that the unrestricted black-box complexity of $\text{ONEMAX}_n$ is $\Theta(n/\log n)$.

The unary unbiased black-box complexity of $\text{ONEMAX}_n$ is $\Theta(n \log n)$ [17]. Higher arity models were studied in [5]. The authors prove that for $2 \le k \le n$ the k-ary unbiased black-box complexity of $\text{ONEMAX}_n$ is $O(n/\log k)$. Theorem 6 shows that we can achieve the same bound in the (much weaker) unbiased ranking-based model.

To ease reading, we split the proof of Theorem 6 into three parts. The first part, Section 4.1, is the easiest. It deals with constant values of $k$. We show that any class of *generalized strictly monotone functions* can be optimized by a ranking-based algorithm in $O(n)$ queries using only binary variation operators. $\text{ONEMAX}_n$ is such a class of generalized strictly monotone functions. For the unary setting, Theorem 6 follows from the facts that (i) already the basic unbiased unary black-box complexity of $\text{ONEMAX}_n$ is $\Omega(n \log n)$ [17] and that (ii) that Random Local Search is a unary unbiased ranking-based algorithm which optimizes $\text{ONEMAX}_n$ in $O(n \log n)$ queries.

The second case, covered in Section 4.2, is the most interesting one. We show that for $k = n$ there exists an unbiased ranking-based algorithm which optimizes every function $\text{OM}_z \in \text{ONEMAX}_n$ using only $O(n/\log n)$ queries. More precisely, we show that after $O(n/\log n)$ samples chosen from $\{0,1\}^n$ independently and uniformly at random, with high probability, it is possible to identify the target string $z$. This random sampling idea is also the basis for the previous results on $\text{ONEMAX}_n$ [3–5, 12]. However, we need to be more careful here as we require all variation operators to be unbiased, and furthermore, other than in all the previously mentioned works, we do not have exact knowledge of the fitness values.

Lastly, we show how to deal with the case of arbitrary $k \in \omega(1), k < n$. We prove that, for any such $k$, we can independently optimize substrings ("blocks") of size $k$ in $O(k/\log k)$ iterations, using only $k$-ary unbiased variation operators. We optimize these blocks sequentially. Since there are $\Theta(n/k)$ such blocks of length $k$, the desired $O(n/\log k)$ bound follows. These are Sections 4.3 and 4.4.

### 4.1 Proof of Theorem 6 for Constant Values $k$

As mentioned above, for $k = 1$ the lower bound in Theorem 6 follows from [17, Theorem 6], which states that the unary unbiased black-box complexity of any class of functions $\{0,1\}^n \to \mathbb{R}$ having a single global optimum is $\Omega(n \log n)$. Clearly, $\text{ONEMAX}_n$ is such a class. Since the $k$-ary unbiased black-box complexity of any class $\mathcal{F}$ of functions is a lower bound for the $k$-ary unbiased ranking-based black-box complexity of $\mathcal{F}$, this also shows that the unary unbiased ranking-based black-box complexity of $\text{ONEMAX}_n$ is $\Omega(n \log n)$.

The upper bound is certified by a simple hill-climber, Random Local Search (cf. Algorithm 5). It is easily verified that the variation operator implicit in the mutation step is a unary unbiased one. The selection depends only on the ranking of the current search point $x$ and its neighbor $y$. Hence, Random Local Search (often abbreviated RLS) is a unary unbiased ranking-based black-box algorithm. By the coupon collector's problem (cf. [20] or any other textbook on randomized algorithms) the expected runtime of RLS on any $\text{ONEMAX}_n$ function

> **Algorithm 5:** Random Local Search for maximizing $f\colon \{0,1\}^n \to \mathbb{R}$.
>
> **1 Initialization:** Sample $x \in \{0,1\}^n$ uniformly at random and query $f(x)$;
> **2 Optimization:** for $t = 1, 2, 3, \ldots$ do
> **3**     Choose $j \in [n]$ uniformly at random;
> **4**     Set $y \leftarrow x \oplus e_j^n$ and query $f(y)$; //mutation step
> **5**     **if** $f(y) \geq f(x)$ **then** $x \leftarrow y$; //selection step is $O(n \log n)$. This concludes our comments on the unary unbiased ranking-based black-box complexity of $\textsc{OneMax}_n$.

For $k \geq 2$ we prove the more general statement that all classes of generalized strictly monotone functions have a binary unbiased ranking-based black-box complexity that is at most linear in $n$ (Lemma 7 below).

Following the standard notation, we write $x \prec y$ if for all $i \in [n]$ we have $x_i \leq y_i$ and if there exists at least one $i \in [n]$ such that $x_i < y_i$. A function $f : \{0,1\}^n \to \mathbb{R}$ is said to be *strictly monotone* if for all $x, y \in \{0,1\}^n$ the relation $x \prec y$ implies $f(x) < f(y)$.

We extend this notation as follows. For all $z \in \{0,1\}^n$ we call $f : \{0,1\}^n \to \mathbb{R}$ *strictly monotone with respect to $z$* if the function $f_z : \{0,1\}^n \to \mathbb{R}, x \mapsto f(x \oplus \bar{z})$ is strictly monotone. As is easy to verify, we have $\arg\max f = z$.

Let $\mathcal{F}$ be a class of real-valued functions defined on $\{0,1\}^n$. We call $\mathcal{F}$ a class of *generalized strictly monotone functions* if for all $f \in \mathcal{F}$ there exists a $z \in \{0,1\}^n$ such that $f$ is strictly monotone with respect to $z$. We note that for all $z \in \{0,1\}^n$ the function $\textsc{Om}_z$ is strictly monotone with respect to $z$.

In [5] the authors argue that the unbiased binary black-box complexity of any class of generalized strictly monotone functions is $O(n)$. Since a formal proof is missing in [5], for completeness, we give a full proof here. It follows closely the ideas presented in [5].

**Lemma 7.** *Let $k \geq 2$ and let $\mathcal{F}$ be a class of generalized strictly monotone functions. The $k$-ary unbiased black-box complexity of $\mathcal{F}$ is at most $4n - 5$.*

For proving Lemma 7 we show that two bit strings $x, y \in \{0,1\}^n$ suffice for encoding which bits have been optimized already. We start with two complementary bit strings $y = \bar{x}$ and throughout the run of the algorithm we ensure that $x_i = y_i$ holds only if the entry $x_i$ in position $i$ is known to equal the entry $z_i$ of the target string. For each bit individually, we test whether it should be set to zero or to one. This yields the linear runtime. We show that all this can be done using binary operators only.

*Proof of Lemma 7.* Note that it suffices to prove the statement for $k = 2$ since for any $\ell > k$ the $\ell$-ary unbiased black-box complexity is bounded from above by the $k$-ary one. We claim that Algorithm 6 certifies Lemma 7.

First we note that the selection/update step (lines 7,8,11) depends only on the rankings of the search points. Therefore, Algorithm 6 is a ranking-based one.

Apart from the uniform sampling variation operator, Algorithm 6 makes use of the following variation operators.

- `complement(·)` is a unary variation operator which, given some $x \in \{0,1\}^n$, returns `complement(x)` $:= \bar{x}$, the bitwise complement of $x$.

- `flipOneWhereDifferent(·,·)` is a binary variation operator which, given some $x, y \in \{0,1\}^n$, first chooses uniformly at random a bit position $j \in \{i \in [n] \mid x_i \neq y_i\}$. With

**Algorithm 6:** A binary unbiased ranking-based black-box algorithm for optimizing generalized strictly monotone functions $f \in \mathcal{F}$.

**1 Initialization:** Sample $x \in \{0,1\}^n$ uniformly at random and query $g(f(x))$;
**2** Set $y \leftarrow \texttt{complement}(x)$ and query $g(f(y))$;
**3 Optimization:** for $t = 1, 2, 3, \ldots$ do
**4**   Sample $w \leftarrow \texttt{flipOneWhereDifferent}(x, y)$ and query $g(f(w))$;
**5**   if $g(f(w)) > g(f(x))$ then
**6**     Set $w' \leftarrow \texttt{dist}_1(x, w)$ and query $g(f(w'))$;
**7**     if $g(f(w')) = g(f(x))$ then Update $x \leftarrow w$; //$x$ and $w$ differ in 1 bit
**8**     else if $g(f(w)) > g(f(y))$ then Update $y \leftarrow w$; //$y$ and $w$ differ in 1 bit
**9**   else if $g(f(w)) > g(f(y))$ then
**10**    Set $w' \leftarrow \texttt{dist}_1(y, w)$ and query $g(f(w'))$;
**11**    if $g(f(w')) = g(f(y))$ then Update $y \leftarrow w$; //$y$ and $w$ differ in 1 bit

  probability $1/2$, it returns $x \oplus e_j^n$ and with probability $1/2$ it returns $y \oplus e_j^n$. That is, with equal probability $\texttt{flipOneWhereDifferent}(x, y)$ either flips exactly one bit in $x$, in which $x$ and $y$ differ, or it flips one such bit in $y$.

- $\texttt{dist}_1(\cdot, \cdot)$ is a binary operator which, given some $x, w \in \{0,1\}^n$ returns $\texttt{dist}_1(x, w) = x$ if the Hamming distance $|x \oplus w|_1$ of $x$ and $w$ equals 1 and it returns $w$ otherwise.

It is easily verified that $\texttt{complement}(\cdot)$ is an unbiased unary variation operator. This follows from the fact that $\overline{x \oplus w} = \bar{x} \oplus w$ for all $x, w \in \{0,1\}^n$ and that $\overline{\sigma(x)} = \sigma(\bar{x})$ for all $\sigma \in S_n$. By a similar reasoning and obeying the fact that the position to be flipped is chosen uniformly, one can easily show that $\texttt{flipOneWhereDifferent}(\cdot, \cdot)$ is unbiased as well. Lastly, we also have $\sigma(\texttt{dist}_1(x, w)) = \texttt{dist}_1(\sigma(x), \sigma(w))$ and $\texttt{dist}_1(x \oplus y, w \oplus y) = \texttt{dist}_1(x, w) \oplus y$ for all $x, y, w \in \{0,1\}^n$ and all $\sigma \in S_n$. This shows that $\texttt{dist}_1(\cdot, \cdot)$ is also unbiased.

For proving that Algorithm 6 indeed certifies Lemma 7, let us fix a function $f \in \mathcal{F}$ and let us assume that $f$ is strictly monotone with respect to some fixed $z \in \{0, 1\}^n$.

We show that throughout the run of Algorithm 6 the following invariant holds: for all $i \in [n]$ we have $x_i = y_i$ only if $x_i = z_i$. After initialization we have $x_i \neq y_i$ for all $i \in [n]$. Hence, by construction, the invariant is satisfied. Once we accept a bit flip of position $i \in [n]$, we necessarily have $x_i = y_i$. Hence, the bit will not be flipped in any further iteration of the algorithm. Furthermore, for any two bit strings $x, w \in \{0, 1\}^n$ with Hamming distance $|x \oplus w|_1 = 1$ we have $f(w) > f(x)$ (and, by definition of $g$, $g(f(w)) > g(f(x))$) if and only if $w_i = z_i$ where $i \in [n]$ is the one position in which $x$ and $w$ differ. This shows that the invariant is always satisfied.

Next we show that Algorithm 6 terminates and that the expected runtime is at most $4n - 5$. First we bound from above the number of queries needed to reduce the Hamming distance of $x$ and $y$ from $n$ to two. To this end, let $x, y \in \{0, 1\}^n$ with $|x - y|_1 > 2$. Then, for $w = \texttt{flipOneWhereDifferent}(x, y)$ either we have $|x \oplus w|_1 = 1$ or $|y \oplus w|_1 = 1$. Both events are equally likely and exactly one of them yields an update of $x$ or $y$, respectively. Therefore, the probability of an update equals $1/2$. If we update any one of the two strings, the Hamming distance of $x$ and $y$ decreases by 1. This implies an expected number of $2(n-2)$ iterations for decreasing the Hamming distance of $x$ and $y$ from $n$ to 2. Any such iteration requires at most two queries, the one for $g(f(w))$ and the one for $g(f(w'))$. Therefore, on average, we need at most $4(n-2)$ queries to reduce the Hamming distance from $x$ and $y$ from $n$ to 2.

Once the Hamming distance of $x$ and $y$ is reduced to two, either we have $z \in \{x, y\}$ or we have that both $|x \oplus z|_1 = 1$ and $|y \oplus z|_1 = 1$. By the random initialization of $x$ and the fact that we replace $x$ and $y$ only by bit strings of Hamming distance one, all bits $x_i$ for which $x_i \neq y_i$ satisfy $\Pr[x_i = 1] = \Pr[x_i = 0] = 1/2$ and, likewise, $\Pr[y_i = 1] = \Pr[y_i = 0] = 1/2$. Therefore, the event $z \in \{x, y\}$ occurs with probability $1/2$. In this case we are done, because both $g(f(x))$ as well as $g(f(y))$ have been queried already. Therefore, let us assume that $z \notin \{x, y\}$. Let again $w = \texttt{flipOneWhereDifferent}(x, y)$. Then both $|x \oplus w|_1 = 1$ and $|y \oplus w|_1 = 1$ must hold. Since only two such strings with Hamming distance one from both $x$ and $y$ exist, we have $\Pr[w = z] = 1/2$. Furthermore, $g(f(w)) > g(f(x))$ or $g(f(w)) > g(f(y))$ holds only if $w = z$. That is, if $w \neq z$, then neither $x$ nor $y$ will be updated. Therefore, once $|x \oplus y|_1 = 2$, it takes, on average, $1/2 \cdot 0 + 1/2 \cdot 2 = 1$ query until Algorithm 6 queries $z = \arg \max f$ for the first time.

Together with the two queries needed for initialization (lines 1 and 2 of Algorithm 6), the runtime of our algorithm can be bounded from above by $2 + 4(n - 2) + 1 = 4n - 5$. □

## 4.2 Proof of Theorem 6 for $k = n$

We show that there exists a ranking-based black-box algorithm which optimizes any function $\text{OM}_z \in \textsc{OneMax}_n$ in $O(n/\log n)$ iterations.

To ease reading of the following description, let us already fix here some unknown function $\text{OM}_z \in \textsc{OneMax}_n$. In order to optimize $\text{OM}_z$, the algorithm has to query the target string $z \in \{0, 1\}^n$.

We work with the monotone model, i.e., whenever the algorithm queries from the oracle a search point $x$, it receives from it the value $g(\text{OM}_z(x))$.

The rough description of the algorithm certifying Theorem 6 for $k = n$ is fairly easy. It first samples $s \in O(n/\log n)$ search points $x_1, \ldots, x_s$ from $\{0, 1\}^n$ mutually independent and uniformly at random. We show that, with high probability, knowing only the $(g \circ \text{OM}_z)$-values $\{g(\text{OM}_z(x_i)) \mid i \in [s]\}$ suffices to create the target string $z$ using only two additional (unbiased) iterations. These last two queries, however, require some technical effort. This is the main part of this section.

In what follows, let $\kappa \geq 2$ be a constant, let $\beta := e^{-4\kappa^2}(2\sqrt{\pi})^{-1}$, and let $\alpha$ be a constant that is at least $8\left(1 - 2e^{-2\kappa^2}\right)^{-1}$. Furthermore, let $s := \alpha n / \ln n$ and let $x_1, \ldots, x_s$ be sampled from $\{0, 1\}^n$ independently and uniformly at random.

We divide the proof of Theorem 6 for the case $k = n$ into three steps. We feel that the intermediate steps are interesting on their own. Each of the following statements holds with exponentially small probability of failure. In particular, they hold with probability at least $1 - o(n^{-\lambda})$ for all constant values of $\lambda$.

- First we show that for each $\ell \in [\frac{n}{2} \pm \kappa\sqrt{n}]$ there is at least one bit string $x_i$ such that $\text{OM}_z(x_i) = \ell$. Furthermore, the set of all samples with $\text{OM}_z$-value in $[\frac{n}{2} \pm \kappa\sqrt{n}]$ has size at least $\frac{s}{2}(1 - 2e^{-2\kappa^2})$.

- In the second part we show how to identify $g(\frac{n}{2})$ and how this knowledge suffices to identify the interval $g([\frac{n}{2} \pm \kappa\sqrt{n}])$. This allows us to calculate $\text{OM}_z(x)$ for all $x$ with $\text{OM}_z(x) \in [\frac{n}{2} \pm \kappa\sqrt{n}]$. That is, for such $x$ we are able to undo the monotone perturbation caused by $g$.

- Lastly, we show that there does not exist any $y \in \{0, 1\}^n \setminus \{z\}$ with $\text{OM}_y(x) = \text{OM}_z(x)$ for all such samples $x \in \{x_1, \ldots, x_s\}$ with $\text{OM}_z(x) \in [\frac{n}{2} \pm \kappa\sqrt{n}]$. Thus, we can unambiguously determine $z$.

**Part 1: Flooding the interval $[\frac{n}{2} \pm \kappa\sqrt{n}]$**

In the next two lemmata we show that, with probability at least $1 - 3\kappa\sqrt{n}\exp(-\frac{\alpha\beta\sqrt{n}}{8\ln n})$, for all $\ell \in [\frac{n}{2} \pm \kappa\sqrt{n}]$ there exists at least one $i \in [s]$ such that $\text{OM}_z(x_i) = \ell$.

**Lemma 8.** *Let $\ell \in [\frac{n}{2} \pm \kappa\sqrt{n}]$ and let $x$ be sampled from $\{0,1\}^n$ uniformly at random. For large enough $n$ we have that $\Pr[\text{OM}_z(x) = \ell] \geq \beta n^{-1/2}$.*

*Proof.* Clearly, $\Pr[\text{OM}_z(x) = \ell] = \binom{n}{\ell} 2^{-n}$. Thus, we have to prove that $\binom{n}{\ell} \geq \beta 2^n n^{-1/2}$, for large enough $n$. Let $\gamma \in [-\kappa, +\kappa]$ such that $\ell = \frac{n}{2} + \gamma\sqrt{n}$.

By definition we have $\binom{n}{\ell} = \frac{n!}{\ell!(n-\ell)!} = \frac{n!}{(n/2+\gamma\sqrt{n})!(n/2-\gamma\sqrt{n})!}$. By Lemma 2 we can bound

$$n! \geq \sqrt{2\pi} n^{n+1/2} e^{-n} e^{1/(12n+1)} \geq \sqrt{2\pi} n^{n+1/2} e^{-n},$$

$$\left(\frac{n}{2} + \gamma\sqrt{n}\right)! \leq \sqrt{2\pi} \left(\frac{n}{2} + \gamma\sqrt{n}\right)^{\frac{n+1}{2}+\gamma\sqrt{n}} e^{-(\frac{n}{2}+\gamma\sqrt{n})} e^{1/(12(\frac{n}{2}+\gamma\sqrt{n}))}, \text{ and}$$

$$\left(\frac{n}{2} - \gamma\sqrt{n}\right)! \leq \sqrt{2\pi} \left(\frac{n}{2} - \gamma\sqrt{n}\right)^{\frac{n+1}{2}-\gamma\sqrt{n}} e^{-(\frac{n}{2}-\gamma\sqrt{n})} e^{1/(12(\frac{n}{2}-\gamma\sqrt{n}))}.$$

We rewrite

$$\left(\frac{n}{2} + \gamma\sqrt{n}\right)^{\frac{n+1}{2}+\gamma\sqrt{n}} = \left(\frac{n}{2}\right)^{\frac{n+1}{2}+\gamma\sqrt{n}} \underbrace{\left(1 + \frac{2\gamma}{\sqrt{n}}\right)^{\gamma\sqrt{n}}}_{(*)} \left(1 + \frac{2\gamma}{\sqrt{n}}\right)^{\frac{n+1}{2}},$$

where term $(*)$ equals $\left((1 + \frac{2\gamma}{\sqrt{n}})^{\frac{\sqrt{n}}{2\gamma}}\right)^{2\gamma^2}$. This term converges to $e^{2\gamma^2}$. For all $n$, term $(*)$ can be bounded from above by $e^{2\gamma^2}$.

Similarly we rewrite

$$\left(\frac{n}{2} - \gamma\sqrt{n}\right)^{\frac{n+1}{2}-\gamma\sqrt{n}} = \left(\frac{n}{2}\right)^{\frac{n+1}{2}-\gamma\sqrt{n}} \underbrace{\left(1 - \frac{2\gamma}{\sqrt{n}}\right)^{-\gamma\sqrt{n}}}_{(**)} \left(1 - \frac{2\gamma}{\sqrt{n}}\right)^{\frac{n+1}{2}},$$

where term $(**)$ equals $\left(\left(1 - \frac{2\gamma}{\sqrt{n}}\right)^{\frac{\sqrt{n}}{2\gamma}}\right)^{-2\gamma^2}$. This term also converges to $e^{2\gamma^2}$. For any $n$, expression $(**)$ can be bounded from below by $e^{2\gamma^2}$. However, by convergence, there exists a $n_0 \in \mathbb{N}$ such that for all $n \geq n_0$ we have $(1 - \frac{2\gamma}{\sqrt{n}})^{-\gamma\sqrt{n}} \leq 2e^{2\gamma^2}$.

Finally, let us note that,

$$\left(1 + \frac{2\gamma}{\sqrt{n}}\right)^{\frac{n+1}{2}} \left(1 - \frac{2\gamma}{\sqrt{n}}\right)^{\frac{n+1}{2}} = \left(1 - \frac{4\gamma^2}{n}\right)^{\frac{n+1}{2}} < 1$$

and, for large enough $n$,

$$e^{1/(12(\frac{n}{2}+\gamma\sqrt{n}))} e^{1/(12(\frac{n}{2}-\gamma\sqrt{n}))} = e^{1/(3n-12\gamma^2)} \leq \sqrt{2}.$$

Altogether we obtain for large enough $n$ that

$$\binom{n}{\ell} \geq \frac{2^{n+1}}{\sqrt{2\pi}\sqrt{n} 2 e^{4\gamma^2}\sqrt{2}} \geq \frac{2^n}{2\sqrt{\pi n}e^{4\kappa^2}} = \beta \frac{2^n}{\sqrt{n}}.$$

□

By applying a Chernoff bound (Lemma 1) to the result of Lemma 8, we immediately obtain the following.

**Corollary 9.** $\Pr\left[\forall \ell \in [\frac{n}{2} \pm \kappa\sqrt{n}] \, \exists i \in [s] : \text{OM}_z(x_i) = \ell\right] \geq 1 - 3\kappa\sqrt{n}\exp(-\frac{\alpha\beta\sqrt{n}}{8\ln n})$.

*Proof.* Throughout this proof assume that $n$ is sufficiently large.

Let $\ell \in [\frac{n}{2} \pm \kappa\sqrt{n}]$. For all $i \in [s]$ let $X_i^\ell$ be the indicator variable of the event $\text{OM}_z(x_i) = \ell$. Let $X^\ell := \sum_{i=1}^s X_i^\ell$. By Lemma 8 we have $\text{E}[X^\ell] \geq s\beta n^{-1/2} = \alpha\beta n^{1/2} \ln^{-1} n$. By applying a Chernoff bound (cf. Lemma 1, (2)) we derive

$$\Pr[X^\ell < \tfrac{\alpha\beta}{2} n^{1/2} \ln^{-1} n] \leq \Pr[X^\ell < \tfrac{1}{2}\text{E}[X^\ell]] \leq \exp(-\tfrac{1}{8}\text{E}[X^\ell]) \leq \exp(-\tfrac{\alpha\beta}{8} n^{1/2} \ln^{-1} n).$$

The statement follows from a simple union bound argument. The probability of not sampling at least one of the values in $[\frac{n}{2} \pm \kappa\sqrt{n}]$ can be bounded from above by $(2\kappa\sqrt{n} + 1)\exp(-\tfrac{\alpha\beta}{8} n^{1/2} \ln^{-1} n)$. □

We conclude the first part by the following elementary lemma, which again is not best possible, but good enough for our purposes. It shows that almost one half of the $s$ samples $x_1, \ldots, x_s$ lie in the interval $[\frac{n}{2} \pm \kappa\sqrt{n}]$.

**Lemma 10.** *(i)* If $x$ is drawn from $\{0,1\}^n$ uniformly at random, then $\Pr[\text{OM}_z(x) \in [\frac{n}{2} \pm \kappa\sqrt{n}]] \geq 1 - 2e^{-2\kappa^2}$.

*(ii)* Let $S$ be the number $|\{i \in [s] \mid \text{OM}_z(x_i) \in [\frac{n}{2} \pm \kappa\sqrt{n}]\}|$ of samples with $\text{OM}_z$-value in $[\frac{n}{2} \pm \kappa\sqrt{n}]$. With probability at least $1 - \exp(-\frac{(1-2e^{-2\kappa^2})\alpha n}{8\ln n})$ we have $S \geq \frac{s}{2}(1 - 2e^{-2\kappa^2})$.

*Proof.* Let $x$ be drawn from $\{0,1\}^n$ uniformly at random. Then, by Chernoff's bound (cf. (1) in Lemma 1),

$$\Pr[\text{OM}_z(x) \in [\tfrac{n}{2} \pm \kappa\sqrt{n}]] = 1 - \Pr[|\text{OM}_z(x) - \text{E}[\text{OM}_z(x)]| > \kappa\sqrt{n}] \geq 1 - 2e^{-2\kappa^2}.$$

This shows **(i)**. Furthermore, we expect at least $(1 - 2e^{-2\kappa^2})s$ samples to have an $\text{OM}_z$-value in $[\frac{n}{2} \pm \kappa\sqrt{n}]$. By again applying a Chernoff bound (cf. (1) in Lemma 1), we bound

$$\Pr[S \leq \tfrac{1}{2}(1 - 2e^{-2\kappa^2})s] \leq \Pr[S \leq \tfrac{1}{2}\text{E}[S]] \leq \exp(-\tfrac{1}{8}\text{E}[S]) = \exp(-\tfrac{1}{8}(1 - 2e^{-2\kappa^2})s).$$

□

**Part 2: Identification of $g(\frac{n}{2})$ and of $g\left([\frac{n}{2} \pm \kappa\sqrt{n}]\right)$**

From the previous part we know that after drawing $s = \alpha n / \ln n$ samples independently and uniformly at random, we can assume that for each value $\ell \in [\frac{n}{2} \pm \kappa\sqrt{n}]$ there exists at least one $i \in [s]$ such that $\text{OM}_z(x_i) = \ell$. Furthermore, we have bounded the number of samples that fall into the interval $[\frac{n}{2} \pm \kappa\sqrt{n}]$. As we shall see in the third part of this section, if we could identify these samples with $\text{OM}_z$-value in $[\frac{n}{2} \pm \kappa\sqrt{n}]$, then, with high probability, we could determine the target string $z$. In this part we show that on top of the $s$ samples $x_1, \ldots, x_s$ we need only one additional query to determine $g(\frac{n}{2})$. Once we have identified the value $g(\frac{n}{2})$, from Part 1 we infer that we also learned $g(\ell)$ for all $\ell \in [\frac{n}{2} \pm \kappa\sqrt{n}]$.

We first explain how to identify $g(\frac{n}{2})$. We do this by exploiting the strong monotonicity of $g$. To be more precise, we make use of the fact that $g$ preserves the element defining the median of a set of objective values. That is, if element $v$ is the median of a multi-set $\{v_1, \ldots, v_t\}$, then $g(v)$ is a median of the multi-set $\{g(v_1), \ldots, g(v_t)\}$. Here in our context we define the median of a finite multi-set $S$ to be the smallest value $v \in S$ such that the number of elements in $S$ which are smaller or equal to $v$ is at least half the size of $S$. Formally, $v = \min\{\ell \in S \mid |\{s \in S \mid s \leq \ell\}| \geq |S|/2\}$. For our context we set

$$m' := \min\left\{\ell \in [0..n] \mid |\{i \in [s] \mid \text{OM}_z(x_i) \leq \ell\}| \geq n/2\right\}.$$

For all statements that follow, we assume that $n$ is large enough.

**Lemma 11.** *The probability that $m' \in [\frac{n}{2} \pm \sqrt{n}]$ is at least $1 - 2\exp(-2\alpha\beta^2 n/\ln n)$.*

*Proof.* We bound the probability that more than $s/2$ samples have an $\text{OM}_z$-value that is less than $\frac{n}{2} - \sqrt{n}$ and we bound the probability that more $s/2$ samples have an $\text{OM}_z$-value that is larger than $\frac{n}{2} + \sqrt{n}$.

Let $x \in \{0,1\}^n$ be sampled uniformly at random. By symmetry, for all $\gamma$ it holds that

$$\Pr[\text{OM}_z(x) = \tfrac{n}{2} + \gamma] = \Pr[\text{OM}_z(x) = \tfrac{n}{2} - \gamma]. \tag{3}$$

Furthermore, by Lemma 8 we have

$$\Pr\left[\text{OM}_z(x) \notin [\tfrac{n}{2} \pm \sqrt{n}]\right] = 1 - \sum_{i=\frac{n}{2}-\sqrt{n}}^{\frac{n}{2}+\sqrt{n}} \Pr[\text{OM}_z(x) = i] \leq 1 - (2\sqrt{n} + 1)\beta n^{-1/2}$$
$$\leq 1 - 2\beta. \tag{4}$$

Equations (3) and (4) imply $\Pr[\text{OM}_z(x) < \tfrac{n}{2} - \sqrt{n}] \leq \tfrac{1}{2} - \beta$. By linearity of expectation we can thus bound the expected number $X$ of samples with $\text{OM}_z$-value less than $\frac{n}{2} - \sqrt{n}$ by $s(\tfrac{1}{2} - \beta)$.

From the Chernoff bound (1) in Lemma 1 we derive

$$\Pr\left[X \geq \tfrac{s}{2}\right] \leq \Pr\left[X \geq \text{E}[X] + s\beta\right] \leq \exp\left(-2s^2\beta^2/s\right) = \exp(-2\alpha\beta^2 n/\ln n)$$

By symmetry, the same reasoning proves $\Pr\left[Y \geq \tfrac{s}{2}\right] \leq \exp(-2\alpha\beta^2 n/\ln n)$ for the number $Y$ of samples with $\text{OM}_z$-value larger than $\frac{n}{2} + \sqrt{n}$. The statement follows from a union bound over the two events $X \geq \tfrac{s}{2}$ and $Y \geq \tfrac{s}{2}$. □

With Lemma 11 at hand, the identification of $g(\tfrac{n}{2})$ and $g\left([\tfrac{n}{2} \pm \kappa\sqrt{n}]\right)$ is easy.

**Lemma 12.** *If we know the $(g \circ \text{OM}_z)$-values of $x_1, \ldots, x_s$, we can apply a unary unbiased variation operator to one of these samples in order to create one additional search point $x'$ such that after querying $g(\text{OM}_z(x'))$ we can identify $g(\lceil \tfrac{n}{2} \rceil)$ and $g\left([\lceil \tfrac{n}{2} \rceil \pm \kappa\sqrt{n}]\right)$, with probability at least $1 - c\sqrt{n}\exp(-\tfrac{\alpha\beta\sqrt{n}}{8\ln n})$ for some constant value $c$.*

*Proof.* Let $m$ be the median of the multi-set $\{g(\text{OM}_z(x_1)), \ldots, g(\text{OM}_z(x_s))\}$. Since $g$ is a strictly monotone function, we have $m = g(m')$. Lemma 11 yields $m \in g\left([\lceil \tfrac{n}{2} \rceil \pm \sqrt{n}]\right)$, with probability at least $1 - 2\exp(-2\alpha\beta^2 n/\ln n)$.

According to Lemma 8, there exists a sample $x_i \in \{x_1, \ldots, x_s\}$ such that $g(\text{OM}_z(x_i)) = m$, with probability at least $1 - 3\kappa\sqrt{n}\exp(-\tfrac{\alpha\beta\sqrt{n}}{8\ln n})$. We show how sampling the bitwise complement $\overline{x_i}$ of $x_i$ reveals $g(\tfrac{n}{2})$.

Before we prove this claim let us first note that $\overline{x_i}$ can be obtained from $x_i$ by the unary unbiased variation operator `complement`$(\cdot)$ which we have introduced in the proof of Lemma 7.

For even values of $n$, our algorithm to identify $g(\tfrac{n}{2})$ is Algorithm 7. Here we denote by

$$\text{median}(g(\text{OM}_z(y)), g(\text{OM}_z(x_i))|g(\text{OM}_z(x_1)), \ldots, g(\text{OM}_z(x_s)))$$

**Algorithm 7:** Identifying $g(\frac{n}{2})$ for even values of $n$.

1. Sample $i \in \{j \in [s] \mid g(\text{Om}_z(x_j)) = m\}$ uniformly at random;
2. Set $y \leftarrow \texttt{complement}(x)$ and query $g(\text{Om}_z(y))$;
3. **if** $g(\text{Om}_z(y)) = g(\text{Om}_z(x_i))$ **then** $m_g \leftarrow m$;
4. **else** $m_g \leftarrow \text{median}(g(\text{Om}_z(y)), g(\text{Om}_z(x_i)) \mid g(\text{Om}_z(x_1)), \ldots, g(\text{Om}_z(x_s)))$;
5. **output** $m_g$

the median of the set (not the multi-set !)

$$\{g(\text{Om}_z(x_1)), \ldots, g(\text{Om}_z(x_s))\} \cap ([g(\text{Om}_z(x_i)), g(\text{Om}_z(y))] \cup [g(\text{Om}_z(y)), g(\text{Om}_z(x_i))]) \,.$$

To show the correctness of Algorithm 7, let us first assume that $g(\text{Om}_z(y)) = g(\text{Om}_z(x_i))$ holds. Since $g$ is a strictly monotone function, this implies $\text{Om}_z(y) = \text{Om}_z(x_i)$. But then $m' = \text{Om}_z(x_i) = \frac{n}{2}$ must hold by the symmetry property that we mentioned already in equation (3) in the proof of Lemma 11.

Therefore, we may assume without loss of generality that $g(\text{Om}_z(x_i)) \neq g(\text{Om}_z(y))$. As mentioned above, by Lemma 11, with probability at least $1 - 2\exp(-2\alpha\beta^2 n/\ln n)$ we have $\text{Om}_z(x_i) = g^{-1}(m) \in [\frac{n}{2} \pm \sqrt{n}]$. So is $g(\text{Om}_z(y))$ by the symmetry of the $\text{Om}_z$ function (equation (3)). The symmetry also implies $\frac{n}{2} = (\text{Om}_z(x_i) + \text{Om}_z(y))/2$.

Assume $\text{Om}_z(x_i) < \text{Om}_z(y)$. Then $\frac{n}{2}$ is exactly the median of the integer values in $[\text{Om}_z(x_i), \text{Om}_z(y)]$. But since we have—by Corollary 9—with probability at least $1 - 3\kappa\sqrt{n}\exp(-\frac{\alpha\beta\sqrt{n}}{8\ln n})$

$$[\text{Om}_z(x_i), \text{Om}_z(y)] \subseteq \{\text{Om}_z(x_1), \ldots, \text{Om}_z(x_s)\},$$

the median of the integer values in $[\text{Om}_z(x_i), \text{Om}_z(y)]$ equals the median of the integer values $\{\text{Om}_z(x_1), \ldots, \text{Om}_z(x_s)\} \cap [\text{Om}_z(x_i), \text{Om}_z(y)]$.

Since $g$ is a strictly monotone function we also have, with the same probability,

$$g\left([\text{Om}_z(x_i), \text{Om}_z(y)]\right) \subseteq \{g(\text{Om}_z(x_1)), \ldots (\text{Om}_z(x_s))\} \,.$$

Therefore, the median of the sampled values

$$\{g(\text{Om}_z(x_1)), \ldots, g(\text{Om}_z(x_s))\} \cap [g(\text{Om}_z(x_i)), g(\text{Om}_z(y))]$$

equals $g(\frac{n}{2})$, if we count each sampled value with multiplicity one.

By Corollary 9, once we have identified $g(\frac{n}{2})$ we also know $g\left([\frac{n}{2} \pm \kappa\sqrt{n}]\right)$, with high probability.

For odd values of $n$ a similar reasoning shows that Algorithm 8 computes $g(\lceil \frac{n}{2} \rceil)$: Either we have $\text{Om}_z(x_i) \in \{\frac{n}{2} - 1, \frac{n}{2} + 1\}$ (lines 3–8) in which case the two values $g(\text{Om}_z(y))$ and $g(\text{Om}_z(x_i))$ must be two consecutive values in $\{g(\text{Om}_z(x_j)) \mid j \in [s]\}$ [1] or we identify as above $g(\lceil \frac{n}{2} \rceil)$ as the median integer value of $[\text{Om}_z(x_i), \text{Om}_z(y)]$ (if $\text{Om}_z(x_i) < \text{Om}_z(y)$) or $[\text{Om}_z(y), \text{Om}_z(x_i)]$ (if $\text{Om}_z(y) < \text{Om}_z(x_i)$), respectively. □

**Part 3: Calculation of $z$**

In this section, we prove that the $s$ random samples and the one additional sample needed to identify $g(\lceil \frac{n}{2} \rceil)$ suffice to determine the target string $z$. We do so by showing that the probability that there exists a bit string $y \neq z$ with $\text{Om}_y(x_i) = \text{Om}_z(x_i)$ for all $i \in [s]$ with $\text{Om}_z(x_i) \in [\frac{n}{2} \pm \kappa\sqrt{n}]$ is small.

---
[1] That is, $\{g(\text{Om}_z(x_1)), \ldots, g(\text{Om}_z(x_s))\} \cap ([g(\text{Om}_z(x_i)), g(\text{Om}_z(y))] \cup [g(\text{Om}_z(y)), g(\text{Om}_z(x_i))]) = \{g(\text{Om}_z(x_i)), g(\text{Om}_z(y))\}$.

**Algorithm 8:** Identifying $g(\lceil \frac{n}{2} \rceil)$ for odd values of $n$.

1 Sample $i \in \{j \in [s] \mid g(\text{OM}_z(x_j)) = m\}$ uniformly at random;
2 Set $y \leftarrow \texttt{complement}(x)$ and query $g(\text{OM}_z(y))$;
3 **if** $g(\text{OM}_z(y)) < g(\text{OM}_z(x_i))$ **then**
4     **if** $\forall j \in [s] : (g(\text{OM}_z(x_j)) \leq g(\text{OM}_z(y))) \vee (g(\text{OM}_z(x_j)) \geq g(\text{OM}_z(x_i)))$ **then**
5         $m_g \leftarrow m$;
6 **else if** $g(\text{OM}_z(y)) > g(\text{OM}_z(x_i))$ **then**
7     **if** $\forall j \in [s] : (g(\text{OM}_z(x_j)) \geq g(\text{OM}_z(y))) \vee (g(\text{OM}_z(x_j)) \leq g(\text{OM}_z(x_i)))$ **then**
8         $m_g \leftarrow g(\text{OM}_z(y))$;
9 **else** $m_g \leftarrow \text{median}(g(\text{OM}_z(y)), g(\text{OM}_z(x_i))|g(\text{OM}_z(x_1)), \ldots, g(\text{OM}_z(x_s))))$;
10 **output** $m_g$

**Lemma 13.** *Let $S := \{i \in [s] \mid \text{OM}_z(x_i) \in [\frac{n}{2} \pm \kappa\sqrt{n}]\}$ be the set of all samples with $\text{OM}_z$-value close to $\frac{n}{2}$. Let $F := \{y \mid \forall i \in S : \text{OM}_z(x_i) = \text{OM}_y(x_i)\}$ be the set of all $y$ that are consistent with the $\text{OM}_z$-values for all $x_i$ with $i \in S$. Let $\lambda$ be a constant.*

*Then, with probability at least $1 - \exp(-\frac{(1-2e^{-2\kappa^2})\alpha n}{8\ln n})$, we have $\mathrm{E}\left[|F|\right] \leq 1 + 2^{-t/4}$ for $t := \frac{s}{2}(1 - 2e^{-2\kappa^2})$.*

*In particular we have that, with probability at least $1 - c\exp(-\frac{(1-2e^{-2\kappa^2})\alpha n}{8\ln n})$ for some constant $c$, there does not exist a string $y \in \{0,1\}^n \setminus \{z\}$ such that $\text{OM}_z(x_i) = \text{OM}_y(x_i)$ for all $i \in S$.*

*Proof.* First note that, by Lemma 10, we can assume that $|S| \geq \frac{s}{2}(1 - 2e^{-2\kappa^2}) = t$, with probability at least $1 - \exp(-\frac{(1-2e^{-2\kappa^2})\alpha n}{8\ln n})$.

Let $y \in \{0,1\}^n \setminus \{z\}$ and let $h := |y \oplus z|_1$ be the Hamming distance of $y$ and $z$. We bound the probability that for all $i \in S$ we have $\text{OM}_y(x_i) = \text{OM}_z(x_i)$.

If we consider one particular sample $x$ chosen from $\{0,1\}^n$ uniformly at random, we have

$$\Pr\left[\text{OM}_y(x) = \text{OM}_z(x) \mid \text{OM}_z(x) \in [\tfrac{n}{2} \pm \kappa\sqrt{n}]\right] \leq \frac{\Pr\left[\text{OM}_y(x) = \text{OM}_z(x)\right]}{\Pr\left[\text{OM}_z(x) \in [\tfrac{n}{2} \pm \kappa\sqrt{n}]\right]}.$$

By Lemma 10 the probability $\Pr\left[\text{OM}_z(x) \in [\tfrac{n}{2} \pm \kappa\sqrt{n}]\right]$ that the $\text{OM}_z$-value of $x$ lies in the interval $[\tfrac{n}{2} \pm \kappa\sqrt{n}]$ can be bounded from below by $1 - 2e^{-2\kappa^2}$.

Furthermore, $\text{OM}_y(x) = \text{OM}_z(x)$ holds if and only if $x$ coincides with $z$ in exactly half of the $h$ bits in which $z$ and $y$ differ. Thus, $\Pr[\text{OM}_y(x) = \text{OM}_z(x)] = \binom{h}{h/2}2^{-h}$ if $h$ is even and $\Pr[\text{OM}_y(x) = \text{OM}_z(x)] = 0$ for odd values of $h$. In particular,

$$\Pr\left[\text{OM}_y(x) = \text{OM}_z(x) \mid x \in S\right] \leq \frac{\binom{h}{h/2}2^{-h}}{1 - 2e^{-2\kappa^2}},$$

for all even values $h$ and $\Pr\left[\text{OM}_y(x) = \text{OM}_z(x) \mid x \in S\right] = 0$ for odd values $h$.

Assume $h$ to be even. As the samples $x_1, \ldots, x_s$ are drawn independently, the probability that $\text{OM}_y(x_i) = \text{OM}_z(x_i)$ for all $i \in S$ can be bounded as follows.

$$\Pr\bigl[\bigwedge_{i\in S}(\text{OM}_y(x_i) = \text{OM}_z(x_i))\bigr] = \prod_{i \in S}\Pr\left[\text{OM}_y(x_i) = \text{OM}_z(x_i)\right] \leq \left(\frac{\binom{h}{h/2}2^{-h}}{1 - 2e^{-2\kappa^2}}\right)^t.$$

As there are $\binom{n}{h}$ different bit strings $y$ with Hamming distance $|y \oplus z|_1 = h$ from $z$, we bound the expected number of bit strings $y \neq z$ with $\text{OM}_y(x_i) = \text{OM}_z(x_i)$ for all $i \in S$ from above by

$$\sum_{h \in [n]; h \text{ even}} \binom{n}{h} \left(\frac{\binom{h}{h/2} 2^{-h}}{1 - 2e^{-2\kappa^2}}\right)^t = \left(1 - 2e^{-2\kappa^2}\right)^{-t} \sum_{h \in [n]; h \text{ even}} \binom{n}{h} \left(\binom{h}{h/2} 2^{-h}\right)^t.$$

It has been proven in [5, Proposition 8] that for sufficiently large $n$ and $\tilde{t} \geq 2\left(1 + \frac{4 \log_2 \log_2 n}{\log_2 n}\right) \frac{n}{\log_2 n}$, it holds that

$$\sum_{h \in [n]; h \text{ even}} \binom{n}{h} \left(\binom{h}{h/2} 2^{-h}\right)^{\tilde{t}} \leq (n/2) \cdot 2^{-3\tilde{t}/4}.$$

In our case the condition $\alpha \geq 8(1 - 2e^{-2\kappa^2})^{-1} > 4(1 - 2e^{-2\kappa^2})^{-1}\left(1 + \frac{4 \log_2 \log_2 n}{\log_2 n}\right)$ ensures that $t$ satisfies this condition. Furthermore, we have for large enough $n$ that $\frac{n}{2} \leq 2^{t/4}$ and lastly, the requirement $\kappa \geq 2$ implies that $2e^{-2\kappa^2} < 0.15$. Hence, $(1 - 2e^{-2\kappa^2})^{-1} \leq 0.85^{-1} \leq 1.18 \leq 2^{1/4}$ and finally, $(1 - 2e^{-2\kappa^2})^{-t} \leq 2^{t/4}$. We thus conclude that, with probability at least $1 - \exp(-\frac{(1 - 2e^{-2\kappa^2})\alpha n}{8 \ln n})$,

$$E[|\{y \neq z \mid \forall x \in S : \text{OM}_y(x) = \text{OM}_z(x)\}|] \leq 2^{-t/4}.$$

□

Before we prove Theorem 6, let us briefly remark the following elementary fact about black-box complexities.

**Lemma 14** (from [6]). *Suppose for an optimization problem $P$ there exists a black-box algorithm $A$ that, with constant success probability, optimizes $P$ in $s$ iterations. Then the black-box complexity of $P$ is at most $O(s)$.*

We are now ready to prove Theorem 6 for the special case of arity $k = n$. To this end, we fix the values of $\kappa := 2$ and $\alpha := 9 = \left\lceil 8(1 - 2e^{-2\kappa^2})^{-1} \right\rceil$.

*Proof of Theorem 6 for $k = n$.* We need to show that there exists a ranking-based unbiased algorithm which optimizes any function $\text{OM}_z \in \text{OneMax}_n$ in an expected number of $O(n/\log n)$ queries.

We claim that Algorithm 9 certifies Theorem 6 for $k = n$ and even values $n$. For odd values, the part in which we identify $g(\lceil \frac{n}{2} \rceil)$ (lines 5–8 of Algorithm 9) needs to be replaced by lines 1–9 of Algorithm 8. We show that the probability that Algorithm 9 queries $z$ after $O(n/\log n)$ iterations is $1 - o(n^{-\lambda})$ for all constant values $\lambda$. By Lemma 14 this implies the desired bound for the $n$-ary black-box complexity of $\text{OneMax}_n$.

First we show that the algorithm employs only unbiased variation operators of arity at most $n$. We have already argued that sampling uniformly at random from $\{0,1\}^n$ is a 0-ary unbiased variation operator and that `complement`$(\cdot)$ is a unary unbiased one. Therefore, we need to show that the operator "sample $x \in F$ uniformly at random" is unbiased and of arity at most $n$. The latter follows from the fact that the size of $S$ is at most $s \in O(n/\log n)$. The unbiasedness of the variation operator follows essentially from the fact that we sample from $F$ uniformly. More

**Algorithm 9:** An $n$-ary unbiased ranking-based black-box algorithm optimizing $\text{ONEMAX}_n$ in $O(n/\log n)$ queries.

1 **Initialization:**
2   **for** $i = 1, \ldots, s$ **do**
3     Sample $x_i \in \{0,1\}^n$ uniformly at random and query $g(\text{OM}_z(x_i))$;
4 **Identification of** $g(\frac{n}{2})$**:**
5   Sample $i \in \{j \in [s] \mid g(\text{OM}_z(x_j)) = m\}$ uniformly at random;
6   Set $y \leftarrow \texttt{complement}(x_i)$ and query $g(\text{OM}_z(y))$;
7   **if** $g(\text{OM}_z(y)) = g(\text{OM}_z(x_i))$ **then** $m_g \leftarrow m$;
8   **else** $m_g \leftarrow \text{median}(g(\text{OM}_z(y)), g(\text{OM}_z(x_i))|g(\text{OM}_z(x_1)), \ldots, g(\text{OM}_z(x_s))))$;
9 Compute $S \leftarrow \{i \in [s] \mid \text{OM}_z(x_i) \in [\frac{n}{2} \pm 2\sqrt{n}]\}$;
10 Compute $F \leftarrow \{y \in \{0,1\}^n \mid \forall i \in S : \text{OM}_z(x_i) = \text{OM}_y(x_i)\}$;
11 Sample $x \in F$ uniformly at random and query $g(\text{OM}_z(x))$;

precisely, let us define the following family of $|S|$-ary distributions over $\{0,1\}^n$. Abbreviate $F(w^1, \ldots, w^{|S|}) := \{y \in \{0,1\}^n \mid \forall i \in [|S|] : \text{OM}_z(w_i) = \text{OM}_y(w_i)\}$ and set

$$D(x|w^1, \ldots, w^{|S|}) := \begin{cases} |F(w^1, \ldots, w^{|S|})|^{-1}, & \text{if } F(w^1, \ldots, w^{|S|}) \neq \emptyset \text{ and } x \in F(w^1, \ldots, w^{|S|}), \\ 0, & \text{if } F(w^1, \ldots, w^{|S|}) \neq \emptyset \text{ and } x \notin F(w^1, \ldots, w^{|S|}) \\ 2^{-n}, & \text{otherwise.} \end{cases}$$

It is now easy to verify that $D(\cdot \mid w^1, \ldots, w^{|S|})_{w^1, \ldots, w^{|S|} \in \{0,1\}^n}$ is a family of unbiased distributions: Let $y \in F(w^1, \ldots, w^{|S|})$ and $v \in \{0,1\}^n$. For all $j \in [|S|]$ clearly we have $\text{OM}_{y \oplus v}(w^j \oplus v) = \text{OM}_y(w^j)$ and, consequently, we have $y \oplus v \in F(w^1 \oplus v, \ldots, w^{|S|} \oplus v)$. Similarly we conclude that for all permutations $\sigma$ of $[n]$ we have $\sigma(y) \in F(\sigma(w^1), \ldots, \sigma(w^{|S|}))$. The same reasoning also proves that $F(w^1, \ldots, w^{|S|}) = \emptyset$ if and only if $F(\sigma(w^1 \oplus v), \ldots, \sigma(w^{|S|} \oplus v)) = \emptyset$.

Each run of Algorithm 9 requires $\alpha n / \ln n + 2 \in O(n/\log n)$ queries. The total probability of failure is at most the sum of the probabilities that

- the median of the $\text{OM}_z(x_i)$-values is not in $[\frac{n}{2} \pm \sqrt{n}]$,

- the probability that there exists a value $\ell \in [\frac{n}{2} \pm 2\sqrt{n}]$ with $\text{OM}_z(x_j) \neq \ell$ for all $j \in [s]$,

- the probability that the size of $S = \{i \in [s] \mid \text{OM}_z(x_i) \in [\frac{n}{2} \pm 2\sqrt{n}]\}$ is less than $\frac{s}{2}(1 - 2e^{-8})$, and

- the probability that in line 11 we do not sample $z$.

Each of these probabilities is at most $o(n^{-\lambda})$ for any constant value $\lambda \in \mathbb{R}$. By a simple union bound we infer that the probability that the target string $z$ is sampled in one run of Algorithm 9 is at least $1 - o(n^{-\lambda})$. This concludes the proof. □

## 4.3 Proof of Theorem 6 for $k \in \Omega(\log^3 n)$

The proof for the general case uses a simple idea also exploited in [5]: Given some arity $k$, $\ln^3 n \leq k < n$, we subdivide the whole bit string into blocks of length $k$. We show that these blocks can be optimized one after the other, each in $O(k/\log k)$ steps. As there are $\lceil n/k \rceil$ such blocks, the desired $O(n/\log k)$ runtime bound follows.

**Algorithm 10:** A $k$-ary unbiased ranking-based black-box algorithm optimizing $\textsc{OneMax}_n$ in $O(n/\log k)$ queries.

1 **Initialization:**
2    Sample $x \in \{0,1\}^n$ uniformly at random and query $g(\textsc{Om}_z(x))$;
3    Set $y \leftarrow \texttt{complement}(x)$ and query $g(\textsc{Om}_z(y))$;
4 **for** $j = 1, \ldots, \lceil n/k \rceil$ **do**
5    Sample $x^{(j,1)} \leftarrow \texttt{flipKWhereDifferent}(x,y)$ and query $g(\textsc{Om}_z(x^{(j,1)}))$;
6    **for** $i = 2, \ldots, s$ **do**
7       Sample $x^{(j,i)}$ from $\{v \in \{0,1\}^n \mid \forall \ell \in I^{(j)} : v_\ell = x_\ell\}$ uniformly at random and query $g(\textsc{Om}_z(x^{(j,i)}))$;
8    Identify $g(\frac{k}{2} + c^{(j)})$; //$c^{(j)}$ is the contribution of bits in $I^{(j)}$ to the $\textsc{Om}_z$-values
9    Compute $S^{(j)}$; //set of samples with $(g \circ \textsc{Om}_z)$-value in $[\frac{k}{2} \pm 2\sqrt{k} + c^{(j)}]$
10   Compute $F^{(j)}$; //set of all feasible bit strings
11   Sample $z^{(j)} \in F^{(j)}$ uniformly at random and query $g(\textsc{Om}_z(z^{(j)}))$;
12   Update $y \leftarrow \texttt{update}(y, x, x^{(j,1)}, z^{(j)})$ and query $g(\textsc{Om}_z(y))$;
13   Update $x \leftarrow z^{(j)}$;

*Proof of Theorem 6 for $k \in \Omega(\log^3 n)$.* To ease reading we assume that $k$ is even. For odd values of $k$, in the following proof all occurrences of $k/2$ must be replaced by $\lceil k/2 \rceil$. Further we skip the "with probability at least..."-statements. Instead, we bound the total failure probability at the end of this proof. Since $k$ is large, a simple union bound will show that we can assume all statements to hold with high enough probability.

Fix $\alpha := 9 = \lceil 8(1 - 2e^{-8})^{-1} \rceil$ and $s := \alpha k / \ln k$. For a better presentation of the ideas let us fix the unknown target function $\textsc{Om}_z \in \textsc{OneMax}_n$.

By Lemma 14 it suffices to show that, with high probability, Algorithm 10 queries the target string $z$. Each run of Algorithm 10 requires $O(n/\log k)$ queries.

The notation used in Algorithm 10 is as follows.

For $x, y \in \{0,1\}^n$ by $O(x, y)$ we denote the set $\{i \in [n] \mid x_i = y_i\}$ of all indices in which $x$ and $y$ coincide. As we shall see below, throughout the run of Algorithm 10, the set $O(x, y)$ equals the set of positions for which we know that $x_i = z_i$ must hold. We call these positions *optimized*.

The variation operator $\texttt{flipKWhereDifferent}(\cdot, \cdot)$ is a binary operator that given two strings $x, y \in \{0, 1\}^n$ picks uniformly at random $k' := \min\{k, n - |O(x,y)|\}$ different elements $i_1, \ldots, i_{k'}$ from the set of positions $[n] \setminus O(x, y)$ in which $x$ and $y$ disagree. It outputs the string $x \oplus e^n_{i_1} \oplus \ldots \oplus e^n_{i_{k'}}$. That is, it flips $k'$ positions in $x$ in which $x$ and $y$ differ. This is easily verified to be an unbiased operator. As in the proof of the unbiasedness of the operator "sample $x \in F$ uniformly at random" in Section 4.2 this follows essentially from the fact that (i) the $k'$ positions are sampled uniformly at random from $[n] \setminus O(x, y)$, that (ii) for all $w \in \{0,1\}^n$ the equation $O(x \oplus w, y \oplus w) = O(x, y)$ holds, and that (iii) for all $\sigma \in S_n$ we have $O(\sigma(x), \sigma(y)) = \sigma(O(x, y))$.

Let $j \in [\lceil n/k \rceil]$. The strings $x$ and $x^{(j,1)}$ are used to encode which substring ("block") is to be optimized in the $j$-th *phase*. Namely, throughout the $j$-th phase all entries in positions

$$I^{(j)} := O(x, x^{(j,1)}) = \{i \in [n] \mid x_i = x_i^{(j,1)}\}$$

remain untouched. We only allow the entries in positions $R^{(j)} := [n] \setminus I^{(j)}$ to be flipped. We call $I^{(j)}$ the set of *irrelevant indices* and we call $R^{(j)}$ the set of *relevant indices*. Unless we are in the very last phase $j = \lceil n/k \rceil$ we have $|I^{(j)}| = n - k$ and $|R^{(j)}| = k$.

Similarly one easily verifies that the operator "Sample $x^{(j,i)}$ from $\{v \in \{0,1\}^n \mid \forall \ell \in I^{(j)} : v_\ell = x_\ell\}$ uniformly at random" employed in line 7 of Algorithm 10 is an unbiased one. The underlying distribution can be specified as

$$D(c \mid a, b) := \begin{cases} 2^{-|a \oplus b|_1}, & \text{if } x_i = a_i \text{ for all } i \in [n] \text{ with } a_i = b_i, \\ 0, & \text{otherwise} \end{cases}$$

for all $a, b, c \in \{0,1\}^n$.

Since we do not touch the entries in positions $I^{(j)}$, all search points sampled in the $j$-th phase (lines 5–13) have an $\text{OM}_z$-value of at least $c^{(j)} := |\{i \in I^{(j)} \mid x_i = z_i\}|$. Our algorithm, of course, does not know the value of $c^{(j)}$. Nevertheless we are able to infer $g(\frac{k}{2} + c^{(j)})$. This can be done as follows. Let $x^{(j,i)}$ be one of the search points $x^{(j,2)}, \ldots, x^{(j,s)}$ sampled in line 7 such that $g(\text{OM}_z(x^{(j,i)}))$ equals the median of the multi-set $\{g(\text{OM}_z(x^{(j,1)})), \ldots, g(\text{OM}_z(x^{(j,s)}))\}$ of the sampled $(g \circ \text{OM}_z)$-values.

Let $\tilde{x}^{(j,i)}$ be the bit string which, on the relevant $k$ bits, is the bitwise complement of $x^{(j,i)}$. Formally, $\tilde{x}_\ell^{(j,i)} := 1 - x_\ell^{(j,i)}$ for all $\ell \in R^{(j)}$ and $\tilde{x}_\ell^{(j,i)} := x_\ell^{(j,i)} = x_\ell$ for all $\ell \in I^{(j)}$. Clearly, $\tilde{x}^{(j,i)}$ can be obtained from $x^{(j,i)}$, $x$, and $x^{(j,1)}$ from the 3-ary unbiased variation operator that, given $a, b, c, d \in \{0,1\}^n$ satisfies

$$D(d \mid a, b, c) = \begin{cases} 1, & \text{if } \forall \ell \in [n] : (b_\ell \neq c_\ell \Rightarrow d_\ell = 1 - a_\ell) \wedge (b_\ell = c_\ell \Rightarrow d_\ell = a_\ell), \\ 0, & \text{otherwise.} \end{cases}$$

To identify $g(\frac{k}{2} + c^{(j)})$, in line 8 we query $g\left(\text{OM}_z(\tilde{x}^{(j,i)})\right)$. Clearly, $g\left(\text{OM}_z(\tilde{x}^{(j,i)})\right) = g\left(\frac{k}{2} + c^{(j)}\right)$ if and only if $g\left(\text{OM}_z(\tilde{x}^{(j,i)})\right) = g\left(\text{OM}_z(x^{(j,i)})\right)$. In case $g\left(\text{OM}_z(\tilde{x}^{(j,i)})\right) \neq g\left(\text{OM}_z(x^{(j,i)})\right)$ we know by Lemma 12 that $g\left(\frac{k}{2} + c^{(j)}\right)$ is the median of the sampled values between $[g\left(\text{OM}_z(\tilde{x}^{(j,i)})\right), g\left(\text{OM}_z(x^{(j,i)})\right)]$ or $[g\left(\text{OM}_z(\tilde{x}^{(j,i)})\right), g\left(\text{OM}_z(x^{(j,i)})\right)]$, respectively, where each sampled $(g \circ \text{OM}_z)$-value is counted with multiplicity one.

Note that once we have determined $g(\frac{k}{2} + c^{(j)})$, we can compute the sets

$$S^{(j)} := \{i \in [s] \mid \text{OM}_z(x^{(j,i)}) \in [\tfrac{k}{2} \pm 2\sqrt{k} + c^{(j)}]\} \text{ and}$$
$$F^{(j)} := \{v \in \{0,1\}^n \mid \left(\forall \ell \in I^{(j)} : v_\ell = x_\ell\right) \wedge \left(\forall i \in S^{(j)} : \text{OM}_z(x^{(j,i)}) = \text{OM}_v(x^{(j,i)})\right)\}.$$

This is due to Lemma 9 where we have shown that for all $\ell \in [\frac{k}{2} \pm 2\sqrt{k}]$ there exists at least one $i$ such that $\text{OM}_z(x^{(j,i)}) = \ell + c^{(j)}$.

We call $F^{(j)}$ the set of all *feasible* bit strings for the $j$-th block. In line 11 we sample from this set uniformly at random. This is an unbiased operation, as can be shown like we did in the proof of Theorem 6 for $k = n$ (see the discussion in that proof just after the definition of Algorithm 9). Note that for identification of $F^{(j)}$ we need at most all samples in $S^{(j)}$ plus the two strings $x$ and $x^{(j,1)}$ which encode the current block we are optimizing. This is, the arity of the corresponding variation operator "sample $x \in F^{(j)}$ uniformly at random" is at most $|S| + 2 \leq 9k/\ln k + 2$. Since we assume that $k \in \Omega(\log^3 n)$, this expression can be bounded by $k$ for large enough $n$.

For the same reason we can assume that $|F^{(j)}| = 1$. This is due to Lemma 13 where we have shown that $|F^{(j)}| = 1$ with probability at least $1 - 2^{-\frac{s}{2}(1 - 2e^{-8})}$. Note that under this assumption, the point $z^{(j)}$ sampled in line 11 coincides with $z$ on all relevant bits $R^{(j)}$ and it coincides with $x$ on all other bits $I^{(j)}$.

In the last step of the $j$-th phase we need to update $x$ and $y$. Recall that by $O(x, y)$ we indicate which bits have been optimized already. So we need to update $O(x, y)$ by adding to it

$R^{(j)}$. This can be done in the following way. First we update $y$ by replacing it with

$$\texttt{update}(y, x, x^{(j,1)}, z^{(j)}) := \begin{cases} z_\ell^{(j)}, & \text{if } \ell \in R^{(j)}, \\ y_\ell, & \text{otherwise.} \end{cases}$$

Formally, $\texttt{update}(\cdot, \cdot, \cdot, \cdot)$ is a 4-ary variation operator that, given some $a, b, c, d \in \{0, 1\}^n$ returns a vector with $(\texttt{update}(a, b, c, d))_i = a_i$ for all $i$ with $b_i = c_i$ and $(\texttt{update}(a, b, c, d))_i = d_i$ for all $i$ with $b_i \neq c_i$. Clearly, $\texttt{update}(\sigma(a \oplus w), \sigma(b \oplus w), \sigma(c \oplus w), \sigma(d \oplus w)) = \sigma(\texttt{update}(a, b, c, d) \oplus w)$ for all bit strings $w \in \{0, 1\}^n$ and all permutations $\sigma \in S_n$. Therefore, $\texttt{update}(\cdot, \cdot, \cdot, \cdot)$ is an unbiased variation operator.

In line 13 we finally update $x$ by replacing it with $z^{(j)}$, i.e., we set $x \leftarrow z^{(j)}$. This concludes the $j$-th phase. Summarizing all the above, we have reduced the Hamming distance from $x$ to $y$ by $k'$. We also have $y_i = x_i = z_i^{(j)}$ for all $i \in R^{(j)}$ and, as we shall see below, with high probability, this translates to $x_i = z_i$ for all $i \in R^{(j)}$.

The total number of search points queried in the $j$-th phase is $s + 3 \in O(k/\log k)$. Hence, the total number of queries made by the algorithm is $\lceil n/k \rceil (s + 3) + 2 \in O(n/\log k)$.

To conclude the proof let us bound the total probability of failure. For each block $j$ the total probability of failure is at most the sum of the probabilities that

- the median of the $\text{OM}_z(x_i)$-values is not in $[\frac{k}{2} \pm \sqrt{k} + c^{(j)}]$,

- the probability that there exists a value $\ell \in [\frac{k}{2} \pm 2\sqrt{k} + c^{(j)}]$ with $\text{OM}_z(x^{(j,i)}) \neq \ell$ for all $i \in [s]$,

- the probability that the size of $S = \{i \in [s] \mid \text{OM}_z(x_i) \in [\frac{k}{2} \pm 2\sqrt{k} + c^{(j)}]\}$ is less than $\frac{s}{2}(1 - 2e^{-8})$, and

- the probability that in line 11 we do not sample $z$.

Each of these probabilities is at most $O(\sqrt{k} \exp(-\sqrt{k}/\log k))$. By the union bound the total failure probability is at most $O(n/k)O(\sqrt{k} \exp(-\sqrt{k}/\log k))$, which due to the fact that $k \in \Omega(\log^3 n)$, is $o(1)$, as desired. □

## 4.4 Proof of Theorem 6 for $k \in O(\log^3 n)$

In the last part of the proof for Theorem 6, we consider here in this section the case $k \in O(\log^3 n)$. Again we do a block-wise optimization of the target function. However, for such small values of $k$, the union bound does not suffice for a high probability statement. Instead, we need to identify ways to ensure that each block-wise optimization yields the desired equality $z_i^{(j)} = z_i$ for all $i \in R^{(j)}$, in the notation of the previous section. This can be done by optimizing the first length-$k$ block using the linear query time strategy implicit in Lemma 7. We use this block as a reference block. By flipping all bits in the reference block and flipping all bits in the block currently under investigation, we can probe whether or not all bits in this block coincide with the corresponding entries of the target string.

*Proof of Theorem 6 for $k \in O(\log^3 n)$.* As we have seen in Section 4.1, for constant values of $k$, Theorem 6 is a special case of Lemma 7. Therefore, we can assume in the following that $k = k(n)$ grows with $n$. Further we assume that $k$ is even. For odd values of $k$, in the following proof all occurrences of $k/2$ must be replaced by $\lceil k/2 \rceil$.

Fix $\alpha := 9 = \lceil 8(1 - 2e^{-8})^{-1} \rceil$ and $s := \alpha k/\ln k$. For a better presentation of the ideas let us fix the unknown target function $\text{OM}_z \in \text{OneMax}_n$.

First we show that Algorithm 11 creates in $k + 2$ queries two search points $x'$ and $y'$ of Hamming distance $n - k$ with the additional property that for all $i \in [n]$ with $x'_i = y'_i$ we know $x'_i = z_i$ with certainty.

---

**Algorithm 11:** A binary unbiased ranking-based black-box algorithm for optimizing the first block of length $k$.

---

**1 Initialization:** Sample $x' \in \{0, 1\}^n$ uniformly at random and query $g(\text{OM}_z(x'))$;
**2** Set $y' \leftarrow \texttt{complement}(x')$ and query $g(\text{OM}_z(y'))$;
**3 Optimization: for** $t = 1, \ldots, k$ **do**
**4**     Sample $w \leftarrow \texttt{flip1WhereDifferent}(x', y')$ and query $g(\text{OM}_z(w))$;
**5**     **if** $g(\text{OM}_z(w)) > g(\text{OM}_z(x'))$ **then** Update $x' \leftarrow w$;
**6**     **else** Update $y' \leftarrow \texttt{update}_2(y', x', w)$;

---

To this end, let us first fix the notation. $\texttt{flip1WhereDifferent}(\cdot, \cdot)$ is the operator introduced in Algorithm 10. It flips exactly one bit value of the first argument. The position is chosen uniformly at random from the set of positions in which the first and the second argument disagree.

$\texttt{update}_2(\cdot, \cdot, \cdot)$ is a variation operator, that creates from $y', x', w \in \{0, 1\}^n$ the string $\texttt{update}_2(y', x', w)$ with $(\texttt{update}_2(y', x', w))_i = y'_i$ if $x'_i = w_i$ and $(\texttt{update}_2(y', x', w))_i = x'_i$ if $x'_i \neq w_i$. This is easily verified to be an unbiased variation operator.

In line 5 of Algorithm 11 either we have $g(\text{OM}_z(w)) > g(\text{OM}_z(x'))$—in which case the position $i$ in which $x'$ and $w$ differ satisfies $x'_i \neq z_i = w_i = y'_i$. In this case, updating $x'$ clearly reduces the Hamming distance of $x'$ and $y'$ by one. It preserves the invariant that $x'_i = z_i$ for all positions $i \in [n]$ with $x'_i = y'_i$. If, alternatively, $g(\text{OM}_z(w)) < g(\text{OM}_z(x'))$ holds, then $x'_i = z_i \neq w_i = y'_i$ and we update $y'$ by replacing its $i$-th bit with $x'_i$. This also reduces the Hamming distance of $x'$ and $y'$ by one, again preserving the invariant $x'_i = y'_i \Rightarrow x'_i = z_i$.

Therefore, after termination of Algorithm 11 we have two bit strings $x', y'$ of Hamming distance $|x' \oplus y'|_1 = k$ that satisfy $(x'_i = y'_i) \Rightarrow (x'_i = z_i)$. In what follows, we call the bit positions $O(x', y') = \{i \in [n] \mid x'_i = y'_i\}$ the "reference block". Next we show how this block allows us to verify that another block of length $k$ is optimized (i.e., that the entries of this block coincide with the entries of the target string $z$).

The basic idea is simple: first we create from $x'$ and $y'$ two strings $x$ and $y$ such that $O(x, y) = O(x', y')$ but $x_i \neq x'_i$ for all $i \in O(x', y')$. Certainly we have $x_i \neq z_i$ for all such $i \in O(x', y')$. Starting from $x$ and $y$, we run the same block-wise optimization routine as in Subsection 4.3 where the blocks are chosen from $[n] \setminus O(x', y')$. If we want to test that $k$ specific bits of some candidate string $z^{(j)}$ coincide with the entries of $z$, all we need to do is to flip in $z^{(j)}$ all $k$ bits of interest as well as all bits in block $O(x', y')$. Flipping the bits in block $O(x', y')$ increases the $\text{OM}_z$-value by $k$ since for all $i \in O(x', y')$, by construction, $z^{(j)}_i = x_i \neq z_i$. Therefore, the $\text{OM}_z$-values of the candidate solution $z^{(j)}$ and its offspring $\tilde{z}^{(j)}$ coincide if and only if $z^{(j)}$ and $z$ coincide in the $k$ bits of interest.

The notation in Algorithm 12 is the same as the one used in Algorithm 10. In addition, we make use of the following operators.

The string $\texttt{initialize}_1(x', y')$ is defined via

$$(\texttt{initialize}_1(x', y'))_i := \begin{cases} 1 - x'_i, & \text{if } i \in O(x', y'), \\ x'_i, & \text{if } i \notin O(x', y'). \end{cases}$$

**Algorithm 12:** A $k$-ary unbiased ranking-based black-box algorithm for $k \in O(\log^3 n) \cap \omega(1)$ optimizing $\textsc{OneMax}_n$ in $O(n/\log k)$ queries.

**1** **Input:** Two bit strings $x'$ and $y'$ with $O(x', y') = k$ and $x'_i = z_i$ for all $i \in O(x', y')$;
**2** **Initialization:**
**3**     Set $x \leftarrow \texttt{initialize}_1(x', y')$ and query $g(\textsc{Om}_z(x))$;
**4**     Set $x \leftarrow \texttt{initialize}_2(x', y')$ and query $g(\textsc{Om}_z(y))$;
**5** **for** $j = 1, \ldots, \lceil (n-k)/k \rceil$ **do**
**6**     **repeat**
**7**       Sample $x^{(j,1)} \leftarrow \texttt{flipKWhereDifferent}(x, y)$ and query $g(\textsc{Om}_z(x^{(j,1)}))$;
**8**       **for** $i = 2, \ldots, s$ **do**
**9**          Sample $x^{(j,i)}$ from $\{v \in \{0,1\}^n \mid \forall \ell \in I^{(j)} : v_\ell = x_\ell\}$ uniformly at random and query $g(\textsc{Om}_z(x^{(j,i)}))$;
**10**       Identify $g(\frac{k}{2} + c^{(j)})$; $//c^{(j)}$ is the contribution of bits in $I^{(j)}$ to the $\textsc{Om}_z$-values
**11**       Compute $S^{(j)}$; //set of samples with $(g \circ \textsc{Om}_z)$-value in $[\frac{k}{2} \pm 2\sqrt{k} + c^{(j)}]$
**12**       Compute $F^{(j)}$; //set of all feasible bit strings
**13**       Sample $z^{(j)} \in F^{(j)}$ uniformly at random and query $g(\textsc{Om}_z(z^{(j)}))$;
**14**       Set $\tilde{z}^{(j)} \leftarrow \texttt{test}(z^{(j)}, x', y', x, x^{(j,1)})$ and query $g(\textsc{Om}_z(\tilde{z}^{(j)}))$;
**15**     **until** $g(\textsc{Om}_z(z^{(j)})) = g(\textsc{Om}_z(\tilde{z}^{(j)}))$;
**16**     Update $y \leftarrow \texttt{update}(y, x, x^{(j,1)}, z^{(j)})$ and query $g(\textsc{Om}_z(y))$;
**17**     Update $x \leftarrow z^{(j)}$;
**18** Set $w \leftarrow \texttt{finish}(x, x', y')$ and query $g(\textsc{Om}_z(w))$;

Similarly, we set

$$(\texttt{initialize}_2(x', y'))_i := \begin{cases} 1 - x'_i, & \text{if } i \in O(x', y'), \\ y'_i, & \text{if } i \notin O(x', y'). \end{cases}$$

The string $\texttt{initialize}_1(x', y')$ is obtained through sampling from the distribution $D(w \mid x', y') = 1$ if $w_i = 1 - x'_i$ if and only if $i \in O(x', y')$. This is an unbiased distribution. Hence, both $\texttt{initialize}_1(\cdot, \cdot)$ and, by similar reasoning, $\texttt{initialize}_2(\cdot, \cdot)$ are unbiased variation operators.

In line 14 we query $\texttt{test}(z^{(j)}, x', y', x, x^{(j,1)})$ which is defined via

$$(\texttt{test}(z^{(j)}, x', y', x, x^{(j,1)}))_i := \begin{cases} 1 - z_i^{(j)}, & \text{if } i \in O(x', y') \text{ or } x_i \neq x_i^{(j,1)}, \\ z_i^{(j)}, & \text{otherwise.} \end{cases}$$

Again this is easily verified to be sampled from an unbiased (5-ary) distribution.

Lastly, we define $\texttt{finish}(x, x')$ via

$$(\texttt{finish}(x, x', y'))_i := \begin{cases} 1 - x_i, & \text{if } i \in O(x', y'), \\ x_i, & \text{if } i \notin O(x', y'). \end{cases}$$

After having optimized all bits in $[n] \setminus O(x', y')$, this operator finally replaces in $x$ the entries in $O(x', y')$ by their complement. Therefore, $\texttt{finish}(x, x', y')$ equals the target string $z$.

Note that Algorithm 12 queries $z$ in line 18 with certainty. Hence, we only need to argue that the expected number of queries of Algorithm 12 is $O(n/\log k)$. We optimize $\lceil (n-k)/k \rceil$ blocks of length $k$. Call each execution of lines 7–14 for optimizing a block $B$ a *run for $B$*. As argued in Section 4.3, for any such block $B$, the probability that in line 14 of Algorithm 12 we have $g(\text{Om}_z(z^{(j)})) = g(\text{Om}_z(\tilde{z}^{(j)}))$ already after the first run for $B$ is at least $1 - o(k^{-\lambda})$ for all constant values $\lambda$. In particular, it is at least constant. This shows that at most a constant number of runs are expected for optimizing any block, compare Lemma 14. Each run requires $s + 2 \in O(k/\log k)$ queries. This shows that we need an expected number of $O(k/\log k)$ queries to optimize any length-$k$ block. Since there are $\lceil (n-k)/k \rceil$ of them, the statement follows by linearity of expectation. □

## 5 The Different Black-Box Complexities of BinaryValue

In the previous section, we have seen that the additional ranking restriction did not increase the black-box complexity of the OneMax functions class. In this section, we show an example where the two kinds of complexities greatly differ. Another simple class of classical test functions does the job, namely the class of generalized binary-value functions.

The binary-value function Bv is defined via $\text{Bv}(x) = \sum_{i=1}^{n} 2^{i-1} x_i$, that is, it assigns to each bit string the value of the binary number it represents. As before, we regard here generalizations of this single function.

In the following we denote by $\delta$ the Kronecker symbol, i.e., for any two numbers $k, \ell \in \mathbb{N}_0$ we have $\delta(k, \ell) = 1$ if $k = \ell$ and $\delta(k, \ell) = 0$ otherwise.

**Definition 15** (BinaryValue function class). *For $z \in \{0,1\}^n$ and $\sigma \in S_n$, we define the function $\text{Bv}_{z,\sigma} : \{0,1\}^n \to \mathbb{N}_0, x \mapsto \text{Bv}(\sigma(x \oplus \bar{z})) = \sum_{i=1}^{n} 2^{i-1} \delta(x_{\sigma(i)}, z_{\sigma(i)})$. We set $\text{Bv}_z := \text{Bv}_{z,\text{id}_{[n]}}$. We define the classes*

$$\text{BinaryValue}_n := \{\text{Bv}_z \mid z \in \{0,1\}^n\},$$
$$\text{BinaryValue}_n^* := \{\text{Bv}_{z,\sigma} \mid z \in \{0,1\}^n, \sigma \in S_n\}.$$

*If $f \in \text{BinaryValue}_n$ ($f \in \text{BinaryValue}_n^*$), there exist exactly one $z \in \{0,1\}^n$ (exactly one $z \in \{0,1\}^n$ and exactly one $\sigma \in S_n$) such that $f = \text{Bv}_z$ ($f = \text{Bv}_{z,\sigma}$). Since $z = \arg\max \text{Bv}_z$ ($z = \arg\max \text{Bv}_{z,\sigma}$), we call $z$ the* target string *of $f$. Similarly, we call $\sigma$ the* target permutation *of $\text{Bv}_{z,\sigma}$.*

We show that the unbiased black-box complexity of the larger class $\text{BinaryValue}_n^*$ is $O(\log n)$, cf. Theorem 16, whereas the unrestricted ranking-based black-box complexity of the smaller class $\text{BinaryValue}_n$ is $\Omega(n)$, cf. Theorem 17.

Let us begin with the upper bound for the unbiased black-box complexity.

**Theorem 16.** *The $*$-ary unbiased black-box complexity of $\text{BinaryValue}_n^*$ (and thus, the one of $\text{BinaryValue}_n$) is at most $\lceil \log_2 n \rceil + 2$.*

For every $z \in \{0,1\}^n$ and $\sigma \in S_n$ the function $\text{Bv}_{z,\sigma}$ has $2^n$ different function values. That is, the function $\text{Bv}_{z,\sigma} : \{0,1\}^n \to [0..2^n - 1]$ is one-to-one. This is in strong contrast to the functions $\text{Om}_z \in \text{OneMax}_n$ which obtain values in $[0..n]$ only and are thus far from being one-to-one functions. Therefore, from each query to a BinaryValue function we obtain much more information about the underlying target string than we would gain from any OneMax function. In particular, for each query $x$ and for each $i \in [n]$ we can derive from $\text{Bv}_{z,\sigma}(x)$ whether or not $x_{\sigma(i)} = z_{\sigma(i)}$. Hence, all we need to do is to identify $\sigma$. This can be done by binary search.

*Proof of Theorem 16.* We show that Algorithm 13 is an unbiased black-box algorithm which optimizes every $\text{Bv}_{z,\sigma} \in \text{BinaryValue}_n^*$ using at most $\lceil \log_2 n \rceil + 2$ queries.

**Algorithm 13:** An unbiased black-box algorithm optimizing BINARYVALUE$_n$ in $\lceil \log_2 n \rceil + 2$ queries.

1. Sample $x^{(1)} \in \{0,1\}^n$ uniformly at random and query $\text{Bv}_{z,\sigma}(x^{(1)})$;
2. **for** $k = 2, \ldots, \lceil \log_2 n \rceil + 1$ **do**
3. $\quad$ Sample $x^{(k)} \leftarrow \texttt{flipHalf}(x^{(1)}, \ldots, x^{(k-1)})$ and query $\text{Bv}_{z,\sigma}(x^{(k)})$;
4. Set $x^{(\lceil \log_2 n \rceil + 2)} \leftarrow \texttt{consistent}(x^{(1)}, \ldots, x^{(\lceil \log_2 n \rceil + 1)})$ and query $\text{Bv}_{z,\sigma}(x^{(\lceil \log_2 n \rceil + 2)})$;

To describe the variation operators used in Algorithm 13, let $k \in \mathbb{N}$ and let $y^{(1)}, \ldots, y^{(k+1)} \in \{0,1\}^n$. Set

$$F^{(1)}(y^{(1)}, y^{(2)}) := \{j \in [n] \mid y_j^{(1)} \neq y_j^{(2)}\} \text{ and}$$
$$F^{(0)}(y^{(1)}, y^{(2)}) := [n] \setminus F^{(1)}(y^{(1)}, y^{(2)}).$$

That is, $F^{(1)}(y^{(1)}, y^{(2)})$ contains exactly those bit positions in which $y^{(1)}$ and $y^{(2)}$ disagree and $F^{(0)}(y^{(1)}, y^{(2)})$ is the set of positions in which $y^{(1)}$ and $y^{(2)}$ coincide.

Let $1 \le \ell < k$. For each $(i_1, \ldots, i_\ell) \in \{0,1\}^\ell$ we set

$$F^{(i_1, \ldots, i_\ell, 1)}(y^{(1)}, \ldots, y^{(\ell+2)}) := \{j \in F^{(i_1, \ldots, i_\ell)}(y^{(1)}, \ldots, y^{(\ell+1)}) \mid y_j^{(\ell+1)} \neq y_j^{(\ell+2)}\} \text{ and}$$
$$F^{(i_1, \ldots, i_\ell, 0)}(y^{(1)}, \ldots, y^{(\ell+2)}) := F^{(i_1, \ldots, i_\ell)}(y^{(1)}, \ldots, y^{(\ell+1)}) \setminus F^{(i_1, \ldots, i_\ell, 1)}(y^{(1)}, \ldots, y^{(\ell+2)}).$$

This way we iteratively define $F^{(i_1, \ldots, i_k)}(y^{(1)}, \ldots, y^{(k+1)})$ for all $(i_1, \ldots, i_k) \in \{0,1\}^k$. For any such vector $(i_1, \ldots, i_k) \in \{0,1\}^k$ the set $F^{(i_1, \ldots, i_k)}(y^{(1)}, \ldots, y^{(k+1)})$ contains exactly the subset of positions from $F^{(i_1, \ldots, i_{k-1})}(y^{(1)}, \ldots, y^{(k)})$ in which $y^{(k)}$ and $y^{(k+1)}$ agree ($i_k = 0$) and for $i_k = 1$ it contains the subset of positions in $F^{(i_1, \ldots, i_{k-1})}(y^{(1)}, \ldots, y^{(k)})$ in which $y^{(k)}$ and $y^{(k+1)}$ disagree.

Let

$$Z(y^{(1)}, \ldots, y^{(\ell)}) := \{y^{\ell+1} \in \{0,1\}^n \mid \forall (i_1, \ldots, i_{\ell-1}) \in \{0,1\}^{\ell-1} :$$
$$|F^{(i_1, \ldots, i_{\ell-1}, 1)}(y^{(1)}, \ldots, y^{(\ell+1)})| = \lfloor |F^{(i_1, \ldots, i_{\ell-1})}(y^{(1)}, \ldots, y^{(\ell)})|/2 \rfloor \}.$$

That is, $Z(y^{(1)}, \ldots, y^{(\ell)})$ is the set of bit strings $y^{(\ell+1)}$, which, for every $(i_1, \ldots, i_{\ell-1})$ partitions the set $F^{(i_1, \ldots, i_{\ell-1})}(y^{(1)}, \ldots, y^{(\ell-1)})$ into two subsets $F^{(i_1, \ldots, i_{\ell-1}, 1)}(y^{(1)}, \ldots, y^{(\ell+1)})$ and $F^{(i_1, \ldots, i_{\ell-1}, 0)}(y^{(1)}, \ldots, y^{(\ell+1)})$ of (almost) equal size.

For all $\ell \in \mathbb{N}$, the variation operator $\texttt{flipHalf}(\cdot, \ldots, \cdot)$ samples from the $\ell$-ary distribution $(D(\cdot | \cdot, \ldots, \cdot))_{y^{(1)}, \ldots, y^{(\ell)} \in \{0,1\}^n}$, which for given $y^{(1)}, \ldots, y^{(\ell)} \in \{0,1\}^n$ assigns to each $y \in \{0,1\}^n$ the probability

$$D(y \mid y^{(1)}, \ldots, y^{(\ell)}) := \begin{cases} |Z(y^{(1)}, \ldots, y^{(\ell)})|^{-1}, & \text{if } y \in Z(y^{(1)}, \ldots, y^{(\ell)}), \\ 0, & \text{otherwise.} \end{cases}$$

This is an unbiased distribution: For all $y, w, y^{(1)}, \ldots, y^{(\ell)} \in \{0,1\}^n$ and all $(i_1, \ldots, i_{\ell-1}) \in \{0,1\}^{\ell-1}$ we have

$$F^{(i_1, \ldots, i_{\ell-1})}(y^{(1)}, \ldots, y^{(\ell)}) = F^{(i_1, \ldots, i_{\ell-1})}(y^{(1)} \oplus w, \ldots, y^{(\ell)} \oplus w).$$

From this we easily obtain that $y \in Z(y^{(1)}, \ldots, y^{(\ell)})$ if and only if $y \oplus w \in Z(y^{(1)} \oplus w, \ldots, y^{(\ell)} \oplus w)$. In addition, for all $\theta \in S_n$ we have

$$F^{(i_1, \ldots, i_{\ell-1})}(\theta(y^{(1)}), \ldots, \theta(y^{(\ell)})) = \theta(F^{(i_1, \ldots, i_{\ell-1})}(y^{(1)}, \ldots, y^{(\ell)})),$$

and thus, $y \in Z(y^{(1)}, \ldots, y^{(\ell)})$ holds if and only if $\theta(y) \in Z(\theta(y^{(1)}), \ldots, \theta(y^{(\ell)}))$. From this and the fact that each bit string in $Z(y^{(1)}, \ldots, y^{(\ell)})$ is assigned the same probability we infer that flipHalf is an unbiased variation operator. This is true for all $\ell$.

For the description of the second variation operator we abbreviate $t := \lceil \log_2 n \rceil + 1$ and we assume that $\mathrm{Bv}_{z,\sigma} \in \text{BinaryValue}_n^*$ is the (unknown) function to be optimized.

For all $y^{(1)}, \ldots, y^{(t)} \in \{0,1\}^n$ let

$$\mathcal{F}^{\text{consistent}}(y^{(1)}, \ldots, y^{(t)}) := \{z' \in \{0,1\}^n \mid \exists \sigma' \in S_n \forall i \in [t] : \mathrm{Bv}_{z',\sigma'}(y^{(i)}) = \mathrm{Bv}_{z,\sigma}(y^{(i)})\},$$

the set of all bit strings that are *consistent* with the queries $y^{(1)}, \ldots, y^{(t)}$. This is the set of all possible target strings. As we shall see below, in line 4 of Algorithm 13 we have $|\mathcal{F}^{\text{consistent}}(x^{(1)}, \ldots, x^{(t)})| = 1$.

Abbreviate $\mathcal{F}^{\text{consistent}}(y^{(1)}, \ldots, y^{(t)}) = \mathcal{F}^{\text{consistent}}$. For $y \in \{0,1\}^n$ set

$$D'(y \mid y^{(1)}, \ldots, y^{(t)}) := \begin{cases} |\mathcal{F}^{\text{consistent}}|^{-1}, & \text{if } y \in \mathcal{F}^{\text{consistent}}, \\ 0, & \text{if } \mathcal{F}^{\text{consistent}} \neq \emptyset \text{ and } y \notin \mathcal{F}^{\text{consistent}}, \\ 2^{-n}, & \text{otherwise.} \end{cases}$$

This is the distribution from which the variation operator consistent$(y^{(1)}, \ldots, y^{(t)})$ samples. It is an unbiased $t$-ary distribution as can be easily verified using the fact that $y \in \mathcal{F}^{\text{consistent}}(y^{(1)}, \ldots, y^{(t)})$ if and only if $y \oplus w \in \mathcal{F}^{\text{consistent}}(y^{(1)} \oplus w, \ldots, y^{(t)} \oplus w)$ and if and only if $\theta(y) \in \mathcal{F}^{\text{consistent}}(\theta(y^{(1)}), \ldots, \theta(y^{(t)}))$ for every $\theta \in S_n$.

In what follows we argue that in line 4 of Algorithm 13 there exists exactly one string $z' \in \mathcal{F}^{\text{consistent}}(x^{(1)}, \ldots, x^{(t)})$. In this case, clearly, $z' = z$ must hold.

Let us first show that from $x^{(1)}, \ldots, x^{(t)}$ we can infer the underlying target permutation $\sigma$. We do so by proving that **(a)** for any $2 \leq k \leq t$ and for all $j \in [n]$ we can determine the index $(i_1, \ldots, i_{k-1}) \in \{0,1\}^{k-1}$ with $\sigma(j) \in F^{(i_1, \ldots, i_{k-1})}(x^{(1)}, \ldots, x^{(k)})$. This suffices to determine $\sigma$ because, by construction, for all vectors $(i_1, \ldots, i_{t-1}) \in \{0,1\}^{t-1}$, we have $|F^{(i_1, \ldots, i_{t-1})}(x^{(1)}, \ldots, x^{(t)})| \leq 1$.

The key argument proving statement **(a)** is the injectivity of $\mathrm{Bv}_{z,\sigma}$, which, for any index $j \in [n]$ and for every search point $x \in \{0,1\}^n$, reveals whether or not $x_{\sigma(j)} = z_{\sigma(j)}$. Let us fix an index $j \in [n]$. Clearly, $\sigma(j) \in F^{(1)}(x^{(1)}, x^{(2)})$ if and only if

- $x^{(1)}_{\sigma(j)} = z_{\sigma(j)}$ and $x^{(2)}_{\sigma(j)} \neq z_{\sigma(j)}$, or
- $x^{(1)}_{\sigma(j)} \neq z_{\sigma(j)}$ and $x^{(2)}_{\sigma(j)} = z_{\sigma(j)}$.

Similarly, if $\sigma(j) \in F^{(i_1, \ldots, i_{k-1})}(x^{(1)}, \ldots, x^{(k)})$ is known, then $\sigma(j) \in F^{(i_1, \ldots, i_{k-1}, 1)}(x^{(1)}, \ldots, x^{(k+1)})$ if and only if

- $x^{(k)}_{\sigma(j)} = z_{\sigma(j)}$ and $x^{(k+1)}_{\sigma(j)} \neq z_{\sigma(j)}$, or
- $x^{(k)}_{\sigma(j)} \neq z_{\sigma(j)}$ and $x^{(k+1)}_{\sigma(j)} = z_{\sigma(j)}$.

Hence, in line 4 of Algorithm 13 we know $\sigma$. Let $z' \in \mathcal{F}^{\text{consistent}}(x^{(1)}, \ldots, x^{(t)})$. By definition, there exists a permutation $\sigma' \in S_n$ such that for all $i \in [t]$ we have $\mathrm{Bv}_{z',\sigma'}(x^{(i)}) = \mathrm{Bv}_{z,\sigma}(x^{(i)})$. Let $j \in [n]$. As we have shown above, we can identify the vector $(i_1, \ldots, i_{t-1})$ such that $\sigma(j) \in F^{(i_1, \ldots, i_{t-1})}(x^{(1)}, \ldots, x^{(t)})$. By construction, also $\sigma'(j) \in F^{(i_1, \ldots, i_{t-1})}(x^{(1)}, \ldots, x^{(t)})$ must hold. This shows $\sigma' = \sigma$. Hence, $\mathrm{Bv}_{z',\sigma'}(x^{(1)}) = \mathrm{Bv}_{z',\sigma}(x^{(1)}) = \mathrm{Bv}_{z,\sigma}(x^{(1)})$. But as this requires $z' = z$, we conclude that indeed $|\mathcal{F}^{\text{consistent}}(x^{(1)}, \ldots, x^{(t)})| = 1$.

Putting everything together we have shown that from the first $t = \lceil \log_2 n \rceil + 1$ samples we can infer both the target permutation $\sigma$ as well as the target string $z$. This can be sampled from an unbiased distribution in the $(\lceil \log_2 n \rceil + 2)$nd query. $\square$

Let us now prove that already the unrestricted ranking-based black-box complexity of BINARYVALUE$_n$ is asymptotically different from the basic unbiased one of BINARYVALUE$_n^*$.

**Theorem 17.** *The unrestricted ranking-based black-box complexity of* BINARYVALUE$_n$ *and* BINARYVALUE$_n^*$ *is larger than* $n - 1$.

As discussed in the introduction, Droste, Jansen, and Wegener [10] implicitly showed a lower bound of $\Omega(n/\log n)$ for the unrestricted ranking-based black-box complexity of BINARYVALUE$_n$. Our lower bound of $n - 1$ is almost tight. An upper bounds of $n + 1$ for the unrestricted ranking-based black-box complexity of BINARYVALUE$_n^*$ can be shown exactly in the same way as in [10, Theorem 5]. Intuitively, the algorithm which starts with a random initial search point and then, from left to right, flips in each iteration exactly one bit shows this bound. This is a deterministic version of Algorithm 6.

For the unbiased black-box complexities of BINARYVALUE$_n$ and BINARYVALUE$_n^*$ the situation is as follows. Both the basic as well as the ranking-based unary unbiased black-box complexity of BINARYVALUE$_n$ are of order $\Theta(n \log n)$. The lower bound follows from the already mentioned theorem in [17, Theorem 6], which implies that any function with a single global optimum has an unary unbiased black-box complexity of $\Omega(n \log n)$. The upper bound follows from the fact that, for example, Random Local Search (Algorithm 5) solves any instance $\text{Bv}_{z,\sigma} \in \text{BINARYVALUE}_n^*$ in an expected number of $O(n \log n)$ queries. The latter follows from the coupon-collector's problem.

For higher arities $k \geq 2$ Theorem 17 and Lemma 7 from Section 4.1 immediately yield the following.

**Corollary 18.** *For all $k \geq 2$, the $k$-ary unbiased ranking-based black-box complexity of* BINARYVALUE$_n$ *and* BINARYVALUE$_n^*$ *is larger than $n - 1$ and it is at most $4n - 5$.*

To derive the lower bound, Theorem 17, we employ Yao's minimax principle [24].

**Theorem 19** (Yao's minimax principle, formulation following [20]). *Let $\Pi$ be a problem with a finite set $\mathcal{I}$ of input instances (of a fixed size) permitting a finite set $\mathcal{A}$ of deterministic algorithms. Let $p$ be a probability distribution over $\mathcal{I}$ and $q$ be a probability distribution over $\mathcal{A}$. Then,*

$$\min_{A \in \mathcal{A}} \mathrm{E}[T(I_p, A)] \leq \max_{I \in \mathcal{I}} \mathrm{E}[T(I, A_q)],$$

*where $I_p$ denotes a random input chosen from $\mathcal{I}$ according to $p$, $A_q$ a random algorithm chosen from $\mathcal{A}$ according to $q$, and $T(I, A)$ denotes the runtime of algorithm $A$ on input $I$.*

We apply Yao's minimax principle in our setting as follows. We show that in the ranking-based black-box model any deterministic algorithm needs an expected number of more than $n - 1$ iterations to optimize $\text{Bv}_z$, if $\text{Bv}_z$ is taken from BINARYVALUE$_n$ uniformly at random. Theorem 19 then implies that for any randomized algorithm $A$ there exist at least one instance $\text{Bv}_z \in \text{BINARYVALUE}_n$ such that it takes, in expectation, at least $n - 1$ iterations for algorithm $A$ to optimize $\text{Bv}_z$. This implies Theorem 17.

The crucial observation is that when optimizing $\text{Bv}_z$ with a ranking-based algorithms, then from $t$ samples we can learn at most $t - 1$ bits of the hidden bit string $z$. This is easy to see for two samples $x, y$. If $\text{Bv}_z(x) > \text{Bv}_z(y)$, we see that $x_k = z_k \neq y_k$, where $k := \max\{j \in [n] \mid x_j \neq y_j\}$, but we cannot infer any information about $x_\ell$ for $\ell \neq k$. Similarly, if we have $t$ samples $x^{(1)}, \ldots, x^{(t)}$ and their corresponding $\mathrm{Bv}_z$-values, we cannot infer $\binom{t}{2}$ bits of information as one might guess, but at most $t-1$ bits. As we shall see, this is an immediate consequence of the following combinatorial lemma.

**Lemma 20.** *Let $t \in [n]$ and let $x^{(1)}, \ldots, x^{(t)}$ be $t$ pairwise different bit strings. For every pair $(i,j) \in [t]^2$ we set $\ell_{i,j} := \max\{k \in [n] \mid x_k^{(i)} \neq x_k^{(j)}\}$, i.e., $\ell_{i,j}$ is the largest bit position in which $x^{(i)}$ and $x^{(j)}$ differ. Then $|\{\ell_{i,j} \mid i,j \in [t]\}| \leq t-1$.*

*Proof.* Let $z \in \{0,1\}^n$. By renaming the bit strings where required, we may assume $\mathrm{Bv}_z(x^{(1)}) > \ldots > \mathrm{Bv}_z(x^{(t)})$.

We prove the statement by induction on $t$. For $t=1$ and $t=2$ there is nothing to show. Therefore, we may assume that we have proven $|\{\ell_{i,j} \mid i,j \in [k]\}| \leq k-1$ for some $k \geq 2$. Now, if $\ell_{h,k+1} \in \{\ell_{i,j} \mid i,j \in [k]\}$ for all $h \in [k]$, then clearly we have $|\{\ell_{i,j} \mid i,j \in [k+1]\}| \leq k-1$. Thus, we may assume without loss of generality that there exists a $h \in [k]$ with $\ell_{h,k+1} \notin \{\ell_{i,j} \mid i,j \in [k]\}$.

Since $\mathrm{Bv}_z(x^{(h)}) > \mathrm{Bv}_z(x^{(k+1)})$, the definition of $\mathrm{Bv}_z$ implies that $z_{\ell_{h,k+1}} = x^{(h)}_{\ell_{h,k+1}} \neq x^{(k+1)}_{\ell_{h,k+1}}$. Now, let $j \in [k]$. We show that either $\ell_{j,k+1} = \ell_{h,k+1}$ or $\ell_{j,k+1} = \ell_{j,h}$.

Let us first consider the case $j \leq h$. Since $\mathrm{Bv}_z(x^{(j)}) > \mathrm{Bv}_z(x^{(h)})$ it holds by the definition of $\mathrm{Bv}_z$ that $z_{\ell_{j,h}} = x^{(j)}_{\ell_{j,h}} \neq x^{(h)}_{\ell_{j,h}}$. Now, either $\ell_{j,h} \geq \ell_{h,k+1}$ or $\ell_{j,h} < \ell_{h,k+1}$. In the first case $x^{(k+1)}_{\ell_{j,h}} = x^{(h)}_{\ell_{j,h}} \neq x^{(j)}_{\ell_{j,h}}$, i.e., $\ell_{j,k+1} \geq \ell_{j,h}$. On the other hand, by definition of the $\ell_{i,i'}$'s, for all $\ell > \ell_{j,h}$ we have $x^{(j)}_\ell = x^{(h)}_\ell$. For the same reason, due to the fact $\ell > \ell_{j,h} \geq \ell_{h,k+1}$, we also have for all such $\ell$ that $x^{(h)}_\ell = x^{(k+1)}_\ell$. From this we infer $\ell_{j,k+1} \leq \ell_{j,h}$. This shows $\ell_{j,k+1} = \ell_{j,h}$.

Equivalently, if $\ell_{j,h} < \ell_{h,k+1}$, then $x^{(j)}_{\ell_{h,k+1}} = x^{(h)}_{\ell_{h,k+1}} \neq x^{(k+1)}_{\ell_{h,k+1}}$. Thus, $\ell_{j,k+1} \geq \ell_{h,k+1}$. On the other hand we have for all $\ell > \ell_{h,k+1} > \ell_{j,h}$ that $x^{(k+1)}_\ell = x^{(h)}_\ell = x^{(j)}_\ell$. This shows $\ell_{j,k+1} \leq \ell_{h,k+1}$ and we conclude $\ell_{j,k+1} = \ell_{h,k+1}$.

The reasoning for $j > h$ is the same. □

We are now ready to prove Theorem 17.

*Proof of Theorem 17.* Since the search space $\{0,1\}^n$ is finite, the set $\mathcal{A}$ of all deterministic algorithms on $\mathrm{BinaryValue}_n$ is finite, if we restrict our attention to those algorithms which stop querying search points after the $n$-th iteration.

As mentioned above, we equip $\mathrm{BinaryValue}_n$ with the uniform distribution. Let $\mathrm{Bv}_z \in \mathrm{BinaryValue}_n$ be drawn uniformly at random and let $A \in \mathcal{A}$ be a (deterministic) algorithm. We show the following statement below.

**(A)** Prior to the $t$-th iteration, the set of still possible target strings has size at least $2^{n-t+1}$ and that all of these target strings have the same probability to be the desired target string.

Consequently, the probability to query the correct bit string in the $t$-th iteration, given that the algorithm has not found it in a previous iteration, is at most $2^{-n+t-1}$. This shows that the expected number of iterations $\mathrm{E}[T(\mathrm{Bv}_z, A)]$ until algorithm $A$ queries the target string $z$ can be bounded from below by

$$\sum_{i=1}^{n} i \cdot \Pr[A \text{ queries } z \text{ in the } i\text{-th iteration}] \geq \sum_{i=1}^{n} i \cdot 2^{-n+i-1} = \sum_{i=1}^{n}(n-i+1)\,2^{-i}$$

$$= (n+1)\sum_{i=1}^{n} 2^{-i} - \sum_{i=1}^{n} i\,2^{-i}. \qquad (5)$$

A simple, but nonetheless very helpful observation shows

$$\sum_{i=1}^{n} i\, 2^{-i} = \sum_{i=1}^{n} 2^{-i} + \sum_{i=1}^{n} (i-1)\, 2^{-i} = (1 - 2^{-n}) + 2^{-1} \sum_{i=1}^{n} (i-1)\, 2^{-(i-1)}$$
$$= (1 - 2^{-n}) + 2^{-1} \sum_{i=1}^{n-1} i\, 2^{-i},$$

yielding $\sum_{i=1}^{n} i\, 2^{-i} = 2(1 - 2^{-n}) - n 2^{-n} = 2 - (n+2) 2^{-n}$.

Plugging this into (5), we obtain

$$\mathrm{E}[T(\mathrm{Bv}_z, A)] \geq (n+1)(1 - 2^{-n}) - (2 - (n+2) 2^{-n}) > n - 1.$$

This proves $\min_{A \in \mathcal{A}} \mathrm{E}[T(\mathrm{Bv}_z, A)] > n - 1$ for $\mathrm{Bv}_z$ taken from $\textsc{BinaryValue}_n$ uniformly at random. Yao's minimax principle implies that for any distribution $q$ over the set of deterministic algorithms we have $\max_{z \in \{0,1\}^n} \mathrm{E}[T(\mathrm{Bv}_z, \tilde{A}_q)] \geq \min_{A \in \mathcal{A}} \mathrm{E}[T(\mathrm{Bv}_z, A)] > n - 1$. That is, the ranking-based black-box complexity of $\textsc{BinaryValue}_n$ is larger than $n - 1$.

It remains to prove **(A)**. Let $t \leq n$ and let $x^{(1)}, \ldots, x^{(t)}$ be the search points which have been queried by the algorithm in the first $t$ iterations. All the algorithm has learned about $x^{(1)}, \ldots, x^{(t)}$ is the ranking of these bit strings induced by $\mathrm{Bv}_z$, i.e., it knows for all $i, j \in [t]$ whether $\mathrm{Bv}_z(x^{(i)}) > \mathrm{Bv}_z(x^{(j)})$, or $\mathrm{Bv}_z(x^{(i)}) < \mathrm{Bv}_z(x^{(j)})$, or $\mathrm{Bv}_z(x^{(i)}) = \mathrm{Bv}_z(x^{(j)})$. Note that $\mathrm{Bv}_z(x^{(i)}) = \mathrm{Bv}_z(x^{(j)})$ implies $x^{(i)} = x^{(j)}$. Thus, this case can be disregarded as one cannot learn any additional information by querying the same bit string twice.

As in Lemma 20 we set $\ell_{i,j} := \max\{k \in [n] \mid x_k^{(i)} \neq x_k^{(j)}\}$ for all $i, j \in [t]$ and we set $\mathcal{L} := \{\ell_{i,j} \mid i, j \in [t]\}$.

Let $\ell \in \mathcal{L}$ and let $i, j \in [t]$ such that $\max\{k \in [n] \mid x_k^{(i)} \neq x_k^{(j)}\} = \ell$. We can fix $z_\ell = x_\ell^{(i)}$ if $\mathrm{Bv}_z(x^{(i)}) > \mathrm{Bv}_z(x^{(j)})$, and we fix $z_\ell = x_\ell^{(j)}$ if $\mathrm{Bv}_z(x^{(i)}) < \mathrm{Bv}_z(x^{(j)})$. That is, we can fix $|\mathcal{L}|$ bits of $z$.

Statement **(A)** follows from observing that for every bit string $z'$ with $z'_\ell = z_\ell$ for all $\ell \in \mathcal{L}$ the function $\mathrm{Bv}_{z'}$ yields exactly the same ranking as $\mathrm{Bv}_z$. Hence, all such $z'$ are possible target strings. Since there is no way to differentiate between them, all of them are equally likely to be the desired target string.

Furthermore, it holds by Lemma 20 that $|\mathcal{L}| \leq t - 1$. This shows that, at the end of the $t$-th iteration, there are at least $2^{n-(t-1)}$ possible target strings. By definition of $\mathcal{L}$, the algorithm has queried at most one of them. Consequently, prior to executing the $(t+1)$-st iteration, there are at most $2^{n-(t-1)} - 1 > 2^{n-t}$ bit strings which are equally likely to be the desired target string. This proves **(A)**. $\square$

Note that already a much simpler proof, also applying Yao's minimax principle, shows the following general lower bound.

**Theorem 21.** *Let $\mathcal{F}$ be a class of functions such that each $f \in \mathcal{F}$ has a unique global optimum and such that for all $z \in \{0,1\}^n$ there exists a function $f_z \in \mathcal{F}$ with $z = \arg\max f_z$. Then the unrestricted ranking-based black-box complexity of $\mathcal{F}$ is $\Omega(n/\log n)$.*

## 6 Conclusions

Motivated by the fact that (i) previous complexity models for randomized search heuristics give unrealistic low complexities and (ii) that many randomized search heuristics only compare objective values, but not regard their absolute values, we added such a restriction to the two existing black-box models. While this does not change the black-box complexity of the OneMax function class (this remains relatively low at $\Theta(n/\log n)$), we do gain an advantage for the BinaryValue function class. Here the complexity is $O(\log n)$ without the ranking restriction, but $\Theta(n)$ in the ranking-based model. Our results thus show that for many (but not all) optimization problems, adding the ranking-basedness condition yields more realistic difficulty estimates than the previous black-box models.

### Acknowledgment

Carola Winzen is a recipient of the Google Europe Fellowship in Randomized Algorithms. This work is supported in part by this Google Fellowship.